\documentclass[lettersize,journal]{IEEEtran}
\usepackage{amsmath,amsfonts}
\usepackage{algorithmic}
\usepackage{algorithm}
\usepackage{array}
\usepackage[caption=false,font=normalsize,labelfont=sf,textfont=sf]{subfig}
\usepackage[numbers]{natbib}
\usepackage{textcomp}
\usepackage{stfloats}
\usepackage{url}
\usepackage{verbatim}
\usepackage{graphicx}
\usepackage{subcaption}
\usepackage{subfig}

\begin{document}

\title{Causal Convolutional Neural Networks as \\ Finite
Impulse Response Filters}

\author{Kiran Bacsa, Wei Liu, Xudong Jian, Huangbin Liang, Eleni Chatzi
\thanks{K. Bacsa, X. Jian and H. Liang are with the Future Resilient Systems, Singapore ETH Centre, Singapore, Singapore.}
\thanks{W. Liu is with the Department of Industrial Systems Engineering and Management, NUS, Singapore, Singapore.}
\thanks{E. Chatzi is with the Department of Civil, Environmental and Geomatic Engineering, ETHZ, Zurich, Switzerland.}
}

{


\maketitle
\IEEEpeerreviewmaketitle

\begin{abstract}
This study investigates the behavior of Causal Convolutional Neural Networks (CNNs) with quasi-linear activation functions when applied to time-series data characterized by multimodal frequency content. 
We demonstrate that, once trained, such networks exhibit properties analogous to Finite Impulse Response (FIR) filters, particularly when the convolutional kernels are of extended length exceeding those typically employed in standard CNN architectures. 
Causal CNNs are shown to capture spectral features both implicitly and explicitly, offering enhanced interpretability for tasks involving dynamic systems. 
Leveraging the associative property of convolution, we further show that the entire network can be reduced to an equivalent single-layer filter resembling an FIR filter optimized via least-squares criteria. 
This equivalence yields new insights into the spectral learning behavior of CNNs trained on signals with sparse frequency content. 
The approach is validated on both simulated beam dynamics and real-world bridge vibration datasets, underlining its relevance for modeling and identifying physical systems governed by dynamic responses.
\end{abstract}


Neural networks have enjoyed wide-spread adoption across various modeling tasks, despite the common pitfall of typically comprising black box models that are often difficult to interpret \cite{Alaa2019}.
It is therefore challenging to tailor a neural network model according to the characteristics of a specific problem: how can we introduce a bias inside a black box?
A common way to introduce biases is through the architecture of the neural network. 
For example, Convolution Neural Networks employ convolutional kernels to force the network to focus on local correlations, which is different from the global connectivity of Multi-Layer Perceptrons. 
This bias is useful for image processing tasks, where the information of a single pixel is highly correlated with its surrounding pixels \cite{Alzubaidi2021:a}. 
For physics-informed neural networks \cite{Karniadakis2021}, the bias to be introduced should reflect the prior knowledge on the physical laws that govern the phenomenon that the model is trying to replicate.
Due to the black box nature of neural networks, such biases need to be implemented explicitly, e.g. with a physics-informed loss function, rather than an implicit bias in the architecture of the model.
In the case of the dynamical behavior of physical systems, a desirable bias should capture the dynamic properties of a system.
Such a bias can typically be found in the frequency domain response, or modal properties of such systems. 
These properties can be then used for design, operation and monitoring tasks, across engineering applications, including the fields of structural health monitoring \cite{Reynders2014:b}, robotics \cite{Bayo1989}, control \cite{Boiko2008}, and molecular dynamics \cite{Strachan2004}.
When excited via an impulse, or when subjected to a white noise excitation, discrete vibrating systems exhibit a more pronounced response around certain natural modes of vibration that are centered around specific spectral lines; a trait which is true for both linear \cite{Reynders2012} and nonlinear dynamical systems \cite{Kerschen2009}. The tracking of these modes in terms of frequency, damping and modal shapes forms the main objective of the so-called Operational Modal Analysis (OMA) task.

Operational Modal Analysis (OMA) encompasses a broad range of techniques aimed at identifying the modal properties, natural frequencies, damping ratios, and the mode shapes of structures under ambient or operational loading conditions \cite{Belouchrani1997, Kerschen2007, Yang2013, Nagarajaiah2017, Reynders:2014a, Shih2011, Jang2012, Nguyen2016, Spiridonakos:2016a, Shokrani2016, Loh2015, YSSD2016, Ding2013, Hester2012, Mcgetrick2012, Aguirre2013}. 
Given its focus on spectral content, OMA naturally motivates the integration of frequency-domain representations in machine learning methods for structural identification. 
Recent advances in this direction include the use of frequency-domain Convolutional Neural Networks (CNNs) for damage localization in nonlinear structural systems \cite{MoralesValdez2020}, frequency-based CNNs for automatic feature extraction and damage detection \cite{He2021}, and graph neural networks trained on spectral data for automated modal identification \cite{Jian:2024a}. 
Other approaches, such as wavelet-integrated CNNs, have been applied to classify damage states in railway bridge structures \cite{Nguyen2024}. 
These efforts highlight a growing recognition of the sparsity and interpretability of structural response data in the frequency domain, which is an aspect that contrasts starkly with the characteristics of image data, for which CNNs were originally designed. 
In image processing, the spatial frequency content typically follows a non-sparse power-law distribution ($1/f^2$), as shown by \citet{VanDerSchaf:1996a}, leading to architectures with an inherent bias toward low-frequency, broadband representations. 
Such biases are not necessarily optimal for structural dynamics tasks, where the key information is often concentrated in narrow frequency bands corresponding to the system's modal behavior.

While frequency-domain analysis is fundamental in modeling dynamical systems \cite{Owen2001, Zhang2022}, its integration within neural networks remains limited and is usually treated as a black-box. 
Recent studies have begun to explore the internal mechanisms of neural architectures, aiming to increase the interpretability of such architectures \cite{Rahaman2019, Jacot2021, Stankovic2023}. 
Notably, \citet{Stankovic2023} showed that trained convolutional neural networks (CNNs) can act as matched filters, while \citet{Rahaman2019} provided evidence of an inherent spectral bias in neural networks, attributing it to the nonlinear characteristics of standard activation functions. 
In this work, we shift the focus from image-based tasks to the analysis of signals generated by dynamical systems, specifically, time-series data capturing system responses. 
Traditional architectures for such data include Recurrent Neural Networks (RNNs) \cite{Rumelhart1987}, Long Short-Term Memory networks (LSTMs) \cite{Hochreiter:1997a}, and more recently, Transformer models \cite{Vaswani2017}. 
These models, while powerful, are inherently nonlinear and rely on complex internal mechanisms (e.g., gating, attention) that obscure physical interpretability. 
This poses a fundamental limitation when addressing inverse problems in engineering, such as system identification \cite{Soderstrom1988}, where the objective is to infer a physically meaningful model from observed time-series data. 
In such cases, the weights and internal representations of nonlinear architectures offer limited transparency, making it difficult to extract interpretable information relevant to the underlying physical system.

To investigate the spectral behavior of neural networks in the context of dynamical systems, we focus on a simplified architecture: a one-dimensional quasi-linear Convolutional Neural Network (CNN). 
Compared to gated recurrent models such as RNNs or LSTMs, CNNs offer a more interpretable structure, particularly when analyzing time-series data. 
By employing linear or quasi-linear activation functions, the network preserves the ability to examine its output in the frequency domain using the Fourier transform, similarly to Linear Time-Invariant (LTI) systems. 
This interpretability comes at a cost: the use of linear activations can exacerbate the vanishing gradient problem in deep networks \cite{Hochreiter:1997a}.
Nonetheless, our findings show that when training such linear CNNs to map white noise inputs to targeted spectral bands, the learned kernels resemble Finite Impulse Response (FIR) filters, consistent with the theoretical predictions of \citet{Gunasekar:2018a}. 
This behavior proves especially valuable for modeling systems subject to broadband excitations where white noise serves as a practical approximation of the input. 
The trained models inherently capture and encode the spectral biases of the data throughout the training process. 
Moreover, we observe that the resulting network behaves similarly to a Least-Squares FIR (LS-FIR) filter, despite no explicit constraints enforcing this form. 
To address the vanishing gradient issue, we apply training strategies such as batch normalization and momentum-based optimization. 
Interestingly, replacing linear activations with hyperbolic tangent (tanh) functions leads to comparable filtering behavior, while reducing approximation error. 
As shown by \citet{DeRyck:2021a}, tanh activations retain universal approximation capabilities and exhibit quasi-linear behavior near zero. 
When the activations of the network lie predominantly in this region, the tanh-based CNN can be interpreted as effectively linear, preserving both interpretability and spectral fidelity.

The contributions of our work are summarized as follows.
We demonstrate that, across a range of simple supervised and unsupervised tasks, linear and tanh-activated CNNs exhibit behavior closely resembling that of Finite Impulse Response (FIR) filters with minimal architectural bias. 
This property enhances the interpretability of neural networks operating on time-series data, effectively ``transparentizing" the models and making them particularly suitable for inverse problems such as system identification. 
In essence, we show that neural networks trained on structural dynamics data perform operations that parallel those of classical structural identification techniques. 
Importantly, our objective is not to replace state-of-the-art methods for modal parameter extraction or mode shape recovery. 
Rather, we reveal that deep learning models, when appropriately configured, inherently replicate the functionality of these established methods, offering a bridge between black-box learning and physically interpretable modeling. 
An overview of the proposed scheme is provided in Figure~\ref{fig:summary}.

\begin{figure*}[h]
\centerline{\includegraphics[width=\textwidth]{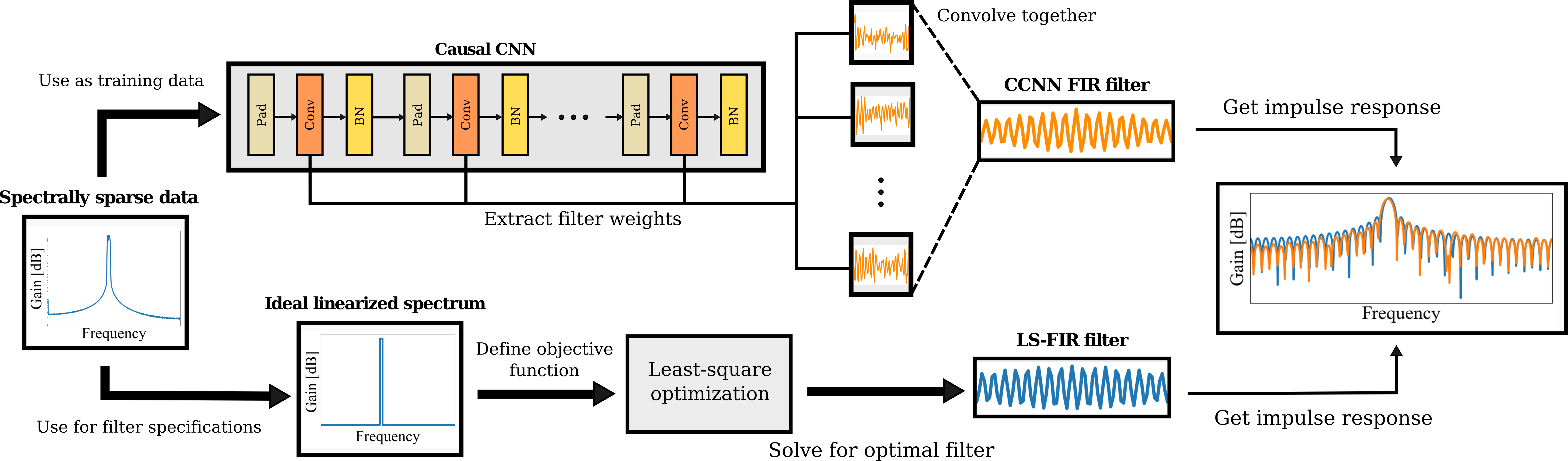}}
\caption{
A spectrally sparse time-series dataset characterized by a few dominant peaks in the frequency domain is used to train a Causal Convolutional Neural Network (CNN). 
After training, the convolutional weights across all layers are combined into an equivalent single-layer filter. 
The resulting impulse response is then compared to that of a Finite Impulse Response (FIR) filter, designed using a least-squares optimization over the same target bandwidth.
}
\label{fig:summary}
\end{figure*}

\section{Relevant works}
\label{relevant}
Fourier theory is basis for many analytical models in signal processing \cite{Kovacevic2013}, and its influence has naturally extended to recent machine learning research. 
A growing body of work has investigated the spectral biases inherent in neural networks, with multiple studies confirming that such models tend to favor low-frequency components during training \cite{Rahaman2019, Zhang2020}. 
One prominent example is the Fourier Neural Operator (FNO) \cite{li2021fourier}, which performs convolution in the frequency domain by applying a Fourier transform over the entire input and retaining a fixed number of modes. 
This approach has achieved state-of-the-art performance in learning operators of partial differential equations (PDEs). 
However, due to the use of nonlinear activations and residual connections, FNOs are not easily interpretable as Linear Time-Invariant (LTI) systems and are thus less suited for inverse problems such as system identification.

Other studies have sought to demystify the inner workings of standard architectures. 
\citet{Stankovic2023}, for example, demonstrated that Convolutional Neural Networks (CNNs) can behave as trainable matched filters, i.e. filters designed to maximize the signal-to-noise ratio by reversing distortions applied to the input signal. 
Although this framework does not explicitly involve Fourier modes, it suggests that CNNs implicitly concentrate on signal components carrying the most energy. 
Given that structural response data exhibits energy concentration in a few dominant spectral modes, training CNNs to minimize output error naturally encourages the kernels to align with these dominant features. 
This aligns well with our interest in spectral sparsity and supports the suitability of CNNs, particularly with linear or quasi-linear activations—for modeling dynamic systems in a physically interpretable manner.

The application of neural networks to system identification has a long history, dating back to early studies in the 1990s that employed Multi-Layer Perceptrons (MLPs) for modeling dynamical systems \cite{Chen1992, Chance1998}. 
More recent efforts have explored recurrent architectures such as Long Short-Term Memory networks (LSTMs), particularly for capturing temporal dependencies in similar settings \cite{Gonzalez2018}. 
In an effort to combine the expressiveness of neural networks with the structure of linear systems, \citet{Piga2021} proposed a neural Linear Time-Invariant (LTI) model designed to learn the transfer function of a dynamical system. 
Their architecture integrates a linear dynamic operator for one-step-ahead prediction, enabling gradient-based training via backpropagation. 
Similarly, \citet{Andersson2019} established a connection between Convolutional Neural Networks (CNNs) and the Volterra series, thereby enabling the use of CNNs for nonlinear system identification.

A particularly notable development is the Structured State Space for Sequence Modeling (S4) architecture introduced by \citet{Gu2022}, which implements a state-space formulation directly within a neural network framework. 
In this model, the state-space matrices are compactly represented as a single causal convolution kernel. 
To reduce computational complexity, the convolution is efficiently implemented as an element-wise product in the Fourier domain. 
Crucially, follow-up work has shown that such linear, structure-aware models not only match but often outperform state-of-the-art deep learning architectures in time-series forecasting tasks \cite{Gu2023}. 
This has reignited interest in linear system modeling within the machine learning community, as evidenced by recent contributions exploring similar directions \cite{Li2024, Zhu2024}. 
Collectively, these works highlight a broader trend toward re-integrating physically grounded, interpretable structures into neural architectures for dynamical system modeling.

On the theoretical front, \citet{Gunasekar:2018a} demonstrated that linear CNNs employing full-width circular convolutions exhibit an inherent sparse spectral bias when trained on risk minimization tasks, such as binary classification. 
While this result offers valuable theoretical insight into the frequency-domain behavior of convolutional networks, its practical implementation remains limited. 
Several challenges hinder its applicability. 
First, full-width convolutions where the kernel spans the entire input are computationally intensive and often infeasible for real-world time-series data. 
Second, although the theoretical framework applies across arbitrary network depth, linear CNNs are known to suffer from vanishing gradient issues, making them difficult to train effectively at scale.

In this work, we build upon the foundational conclusions of \citet{Gunasekar:2018a} and propose a practical architecture for extracting spectrally sparse features from structural dynamic data. 
To this end, we replace full-width circular convolutions with left-padded, long convolutions preserving the temporal causality required for time-series modeling while significantly reducing computational burden. 
Additionally, we substitute purely linear activations with hyperbolic tangent functions to form a quasi-linear CNN. 
This modification mitigates the vanishing gradient problem while preserving near-linear behavior in the vicinity of zero, which is an operating regime our trained network tends to remain within. 
Importantly, we show that the observed spectral sparsity in the learned filters arises primarily from the increased kernel length, rather than being a direct consequence of an implicit bias induced by gradient descent. 
This adaptation renders the spectral insights of \citet{Gunasekar:2018a} practically usable for engineering applications, such as system identification in structural dynamics.

\section{Signal processing and deep learning}

\subsection{Argument for longer kernels}
\label{arguments}

\subsubsection{Finite Impulse Response filters}
Finite Impulse Response filters, as the name suggests, have an impulse response that is finite, i.e., that decays to zero within a finite time.
This is because unlike most other types of filters, the FIR filters do not incorporate feedback mechanisms. 
For a filter of length $p$, the output discrete time series $\mathbf{y}$, generated by convolving the input $\mathbf{x}$ with weights $w_0, ..., w_{p-1}$, is defined as:

\begin{equation}
y[n] = w_{p-1} x[n] + ... + w_0 x[n-p] = \sum_{i=0}^{p-1} w_{p-i} x[n-i]
\end{equation}

This corresponds to a discrete convolution of the input with the filter weights over $i = 1,...,N$ time steps. 
A key advantage of FIR filters is that they are causal and free of poles, which ensures stability. 
This property facilitates the optimization of filter weights using gradient descent methods.

\begin{figure}[H]
\centering
\includegraphics[width=0.7\columnwidth]{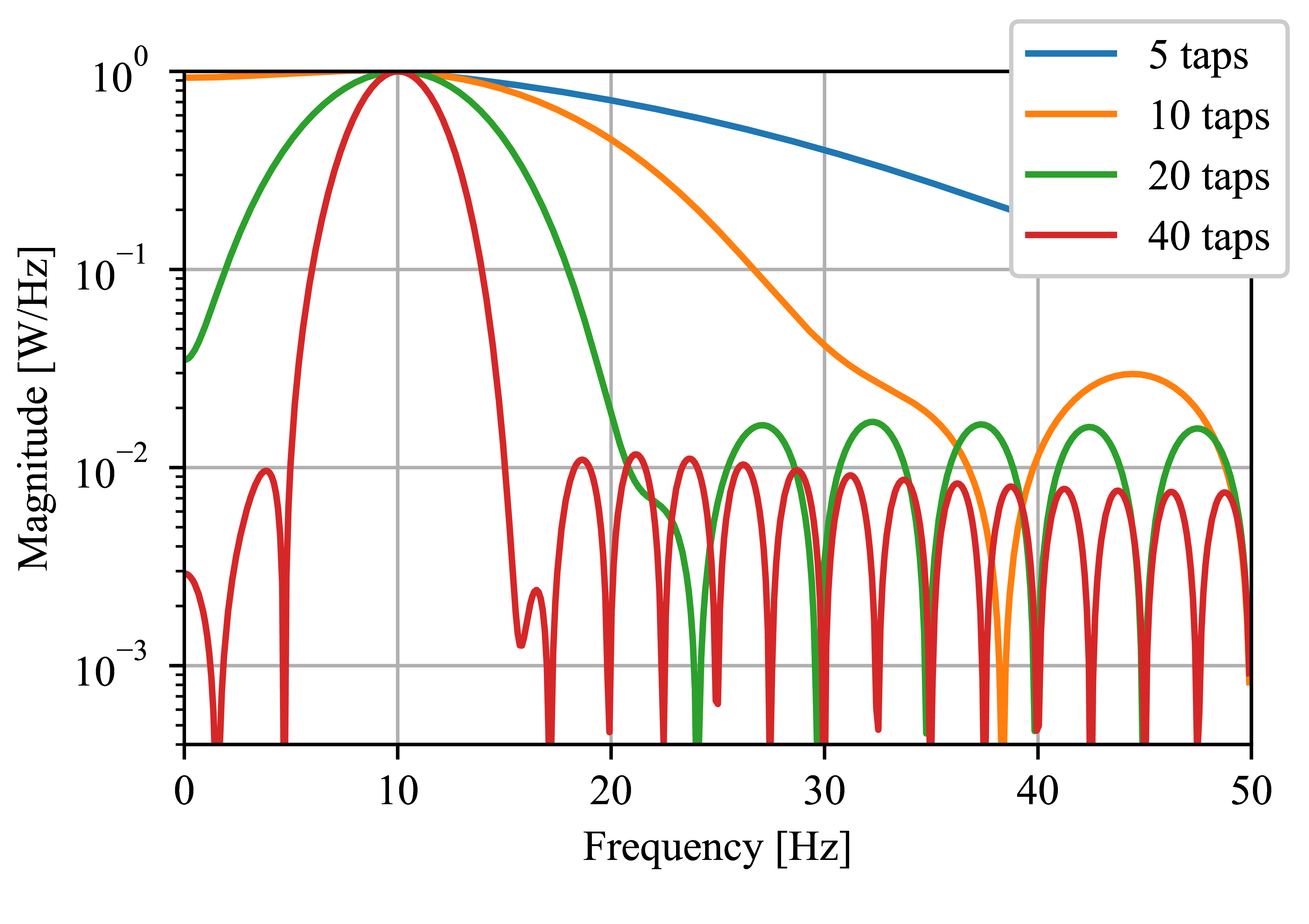}
\caption{
Band-pass FIR filters with a peak at 10 Hz.
Shorter filters are more akin to low-pass than band-pass filters.
The FIR filters are designed using the windowing method, adopting a Hamming window \citep{Parks1972}.
}
\label{fir-filters-lengths}
\end{figure}

\subsubsection{Convolutional Neural Networks}
Convolutional neural networks are a class of feed-forward networks where the forward pass is defined via a convolutional operations, analogous to FIR filters. 
To mimic a causal FIR filter, the convolution kernel must be aligned with the input signal. 
This is typically achieved by padding the signal with $p$ zeros at the beginning, an approach known as ``left-padding.” 
In this setup, the network output follows:

\begin{gather} 
\begin{split}
y[0] = w_{p-1} x[0] \\ 
y[1] = w_{p_1} y[1] + w_{p-2} y[0] \\ 
... \\
y[p] = \sum_{i=0}^{p-1} w_{p-i} x[p-i]
\end{split}
\end{gather}

\subsubsection{Autoregressive models}

Autoregressive (AR) models are commonly used for time-series modeling and describe the evolution of a stochastic variable based on its own prior values \citep{Bendat2010}. 
In an $AR(p)$ model, the current value $\mathbf{x}_t$ is estimated as a linear combination of its previous $p$ values, plus a stochastic noise term $\epsilon_t$:

\begin{equation}
\mathbf{x}_t = \sum_{i=0}^{p-1} w_{p-i} \mathbf{x}_{t-i} + \epsilon_t
\end{equation}

Here, $w_0, ..., w_{p-1}$ are the model parameters and $\epsilon_t$ represents white noise. 
An AR model can be interpreted as the output of an FIR filter driven by white noise, assuming the model is stable and causal. 
Given the variance $Var(\mathbf{x}t) = \sigma{\mathbf{x}}^2$, the spectral density $S(f)$ at frequency $f$ is given by:

\begin{equation}
S(f) = \frac{\sigma_{\mathbf{x}}^2}{|1 - \sum_{j=0}^{p-1} w_{p-j} e^{- (j+1) 2 \pi i f}|^2}
\end{equation}

Low-order AR models typically exhibit heavy spectral tails or peaks at the spectrum boundaries, resembling low-pass or high-pass filters. 
These spectral characteristics, often referred to as red or blue noise, indicate a bias toward low or high frequencies, respectively. 
To isolate narrow-band features, such as modal peaks, for operating engineered systems, higher-order models are often required, often with $p > 20$.

This observation is particularly relevant given that most CNNs reported in the literature use kernel sizes with $p < 10$, effectively biasing them toward low- or high-pass filtering behavior, regardless of the sampling rate. 
This preference for short kernels likely stems from early CNN applications in image processing, where local spatial correlations dominate and spectral resolution is less critical. 
However, this design bias has carried over into time-series tasks, including those related to dynamic systems. 
In this chapter, we advocate for the use of longer convolutional kernels in CNNs to enable more selective filtering in the frequency domain. 
By increasing kernel length, we aim to target specific spectral peaks, such as modal resonances commonly observed in structural dynamic data.


\section{Methodology}
\label{methodology}

We consider a simple model consisting of a series of CNN layers with hyperbolic tangent activations, without added bias, and with batch normalization. 
To ensure causality, we zero-pad the input signal by the length of the convolutional kernel, thereby forcing the network to be autoregressive.
We employ kernel lengths that are significantly longer than those commonly used in existing literature.
While standard convolutional kernels typically span 1 to 7 lags, we advocate using lengths exceeding 35, as argued in Section~\ref{arguments}.
For continuous time FIR filters, only with such a length is it possible to obtain narrow band-pass filters.
We illustrate our argument in Figure \ref{fir-filters-lengths}.
We train our model using the MSE, given as:

\begin{equation}
\mathcal{L}_{\mathrm{MSE}}(\mathbf{x}, \mathbf{y}) = ||f(\mathbf{x}) - \mathbf{y}||_2^2
\end{equation}

Furthermore, since we want to compare our filter with hand-crafted filters with linear phases, we use the associative properties of convolution so that we can combine multiple filters by convolving these together.
For $M$ filters $\{w^i\}_{i=1...M}$, the final filter is given by:

\begin{equation}
W = \mathbf{w}^1 * \mathbf{w}^2 * ... * \mathbf{w}^M
\end{equation}

Where $*$ denotes the convolution operation. 
We impose a symmetric loss regularization on the weights of this final filter so as to impose a linear phase on the spectrum of the learned filter.
In signal processing, phase linearity is a desirable property for filters, as this limits phase distortions of a signal. 
In our work, this step allows us to constrain the solution space of our optimization problem to achieve better convergence.
This regularization loss is given by:

\begin{equation}
\mathcal{L}_{\mathrm{sym}}(W) = ||W - W^{\mathrm{T}}||_2^2
\end{equation}

The final training loss results as the sum of these losses:

\begin{equation}
\mathcal{L}_{\mathrm{tot}} = \mathcal{L}_{\mathrm{MSE}} + \mathcal{L}_{\mathrm{sym}}
\end{equation}

\section{Simulations}
For all simulations, we use the following training parameters.
The Causal CNN (CCNN) is trained using the Adam optimizer from the Pytorch package \citep{pytorch} with a starting learning rate of $10^{-3}$ that is exponentially decayed to $10^{-5}$.
All weights are initialized using Xavier initialization \cite{Glorot2010}.
Manual filters and dynamical simulations are implemented using the SciPy package \cite{scipy2020}.
Simulations were run on an Intel Xeon W-2125 @ 4GHz CPU and an NVIDIA Quadro P400 GPU.
Computation time varies with model complexity: approximately 5 minutes for Section~\ref{target-spectrum}, 1 hour for Section~\ref{regression}, and 3 hours for Section~\ref{unsupervised}.

\label{simulations}

\subsection{Spectrum learning}
\label{target-spectrum}

We first evaluate whether a CCNN trained via gradient descent can approximate a band-pass FIR filter.
To this end, we construct a simple dataset as follows.
We therefore create a simple task with the following dataset.
Random input vectors $\mathbf{x}$ are generated from white noise and then passed through a Chebyshev Type II band-pass filter (order 8, 80 dB attenuation) with a single peak centered at either 10, 23, or 36 Hz, and a bandwidth of 1 Hz. 
The sampling frequency is fixed at 100 Hz.
These filtered outputs serve as the target signals $\mathbf{y}$ for CCNN training.
We train our model on 9000 such input-output pairs, with the goal of reconstructing the filtered output $\mathbf{y}$ from the raw noise input $\mathbf{x}$.
In \ref{hyperparameters}, we discuss the sensitivity of the model to hyperparameters.
Notably, a single-layer CCNN fails to achieve satisfactory results, whereas a 3-layer CCNN successfully isolates the target spectral band from white noise, as shown in Figure~\ref{fig:comparison-layers}.

\begin{figure}[H]
\centerline{\includegraphics[width=0.6\columnwidth]{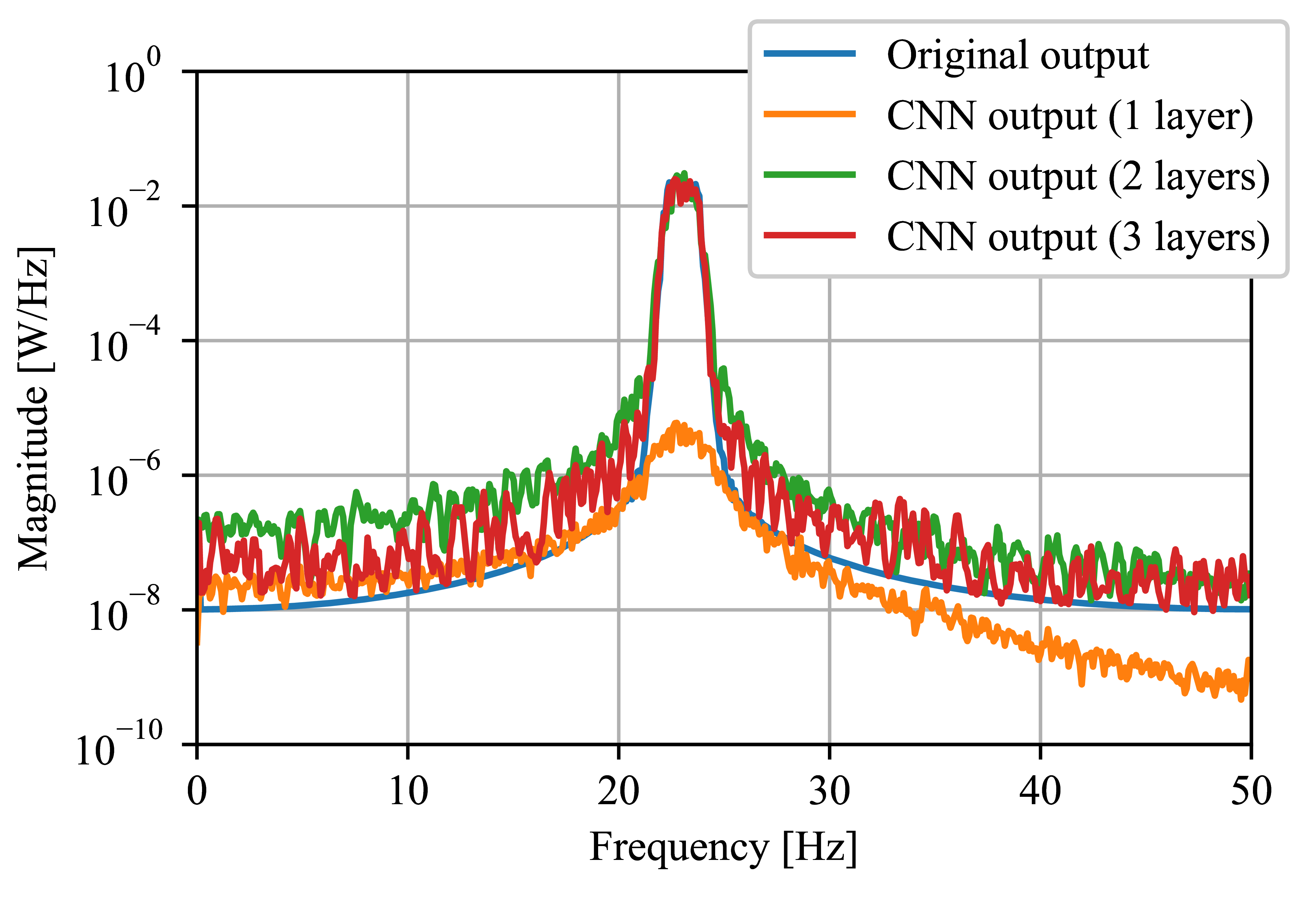}}
\caption{
Comparison of the spectrum of different filters (with 51 taps) against the reference filter spectrum.
}
\label{fig:comparison-layers}
\end{figure}
Finally, we compare the weights and frequency response of the final learned filter, $W$, with those of a filter designed with the window method \citep{Parks1972}.
Once the model is trained, the $R^2$ coefficient of determination for each layer, used to measure linearity, averages $0.975$.
This high score allows us to interpret the CCNNs as quasi-linear, thereby justifying the combination of weights across layers into a single equivalent filter, as in the case of strictly linear activations.
The results are shown for different band-pass peaks in \ref{comparison-10}, \ref{comparison-23},\ref{comparison-36}.
\begin{figure}[H]
\centerline{\includegraphics[width=0.8\columnwidth]{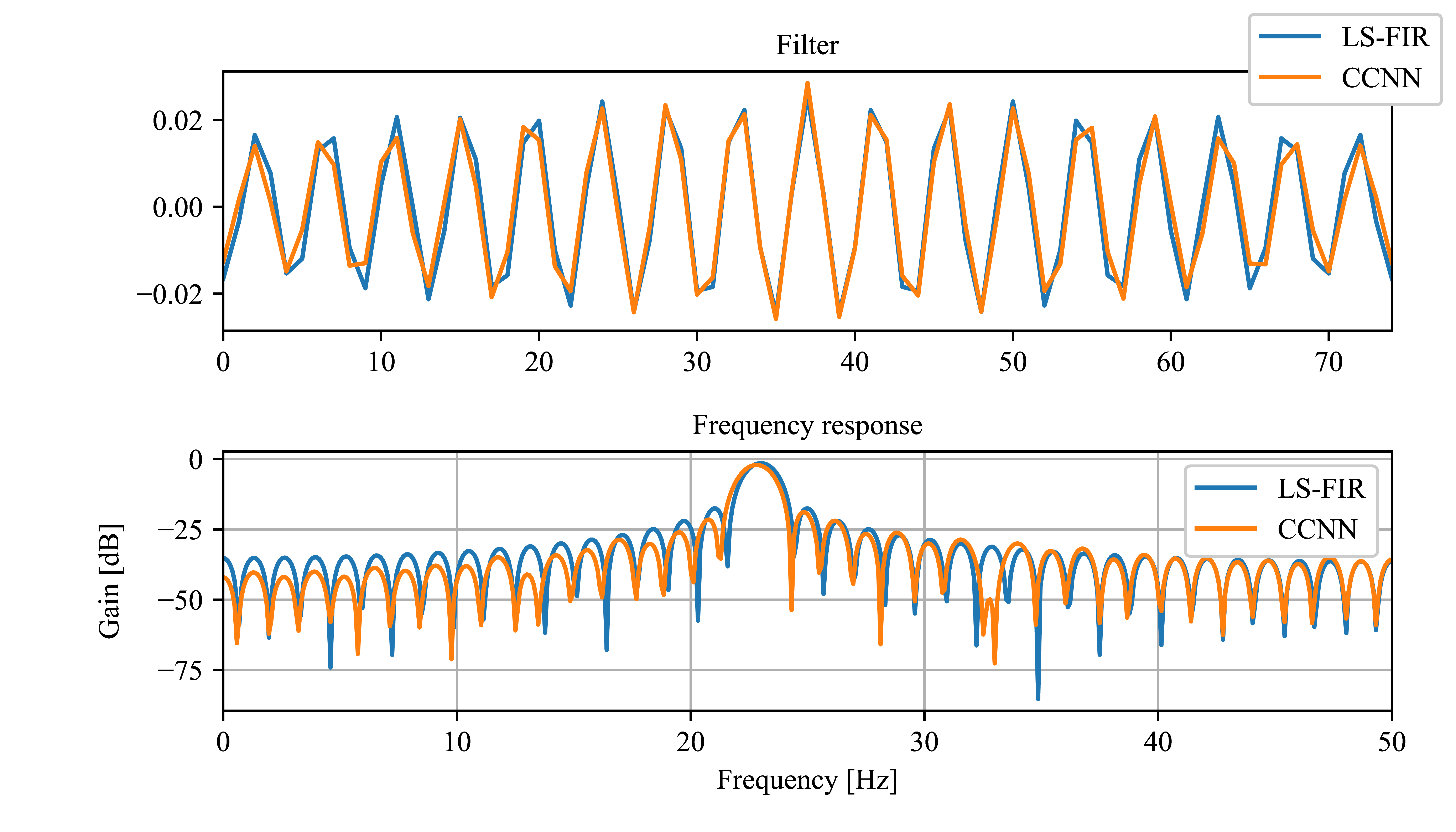}}
\caption{
A comparison of the weights and spectrum the LS-FIR and CCNN filters for a band-pass peak at 23 Hz.
The CCNN here is composed of 7 layers with a kernel width of 75.
}
\label{comparison-23}
\end{figure}
\begin{figure}[H]
\centerline{\includegraphics[width=0.8\columnwidth]{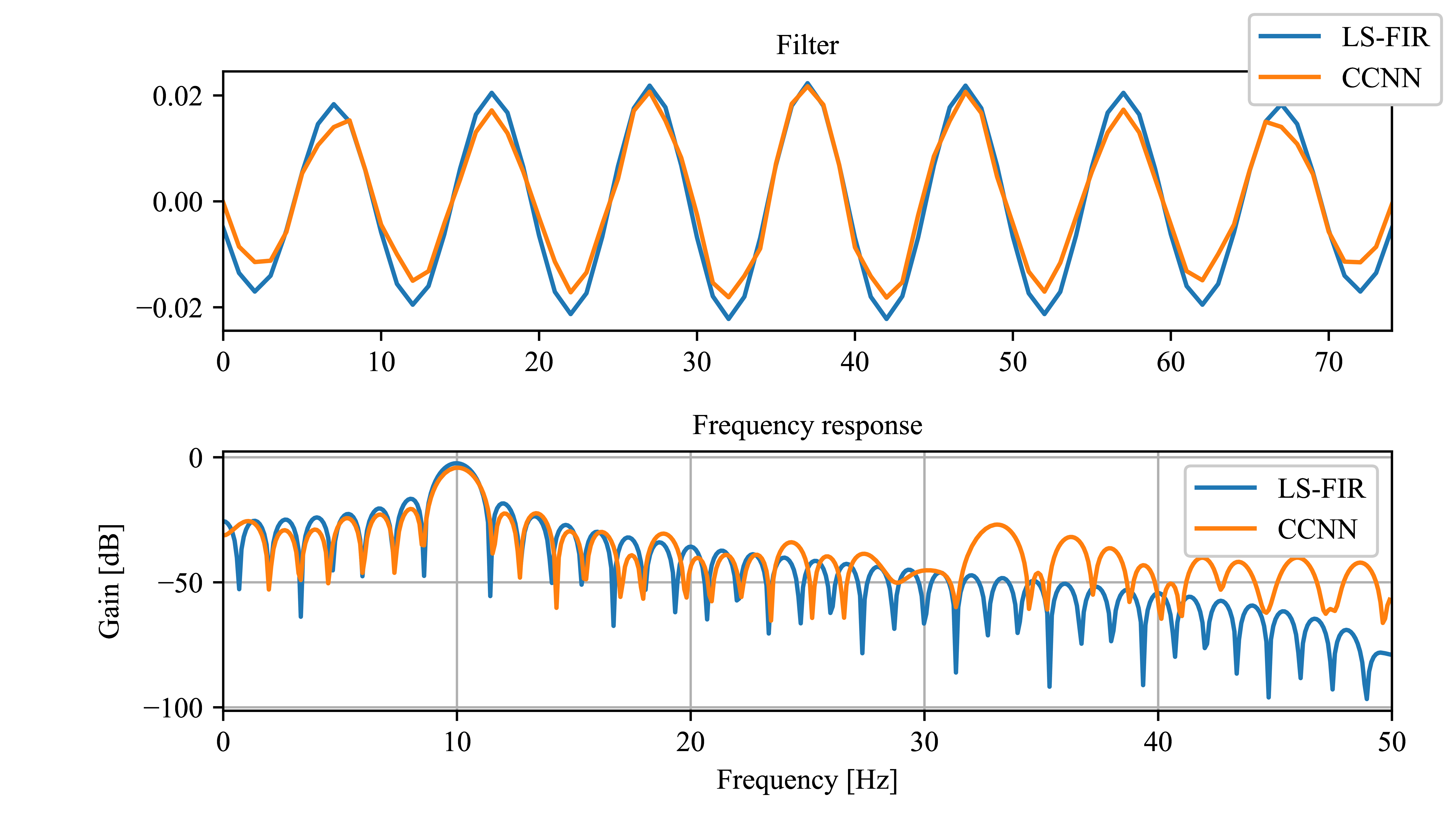}}
\caption{
A comparison of the weights and spectrum the LS-FIR and CCNN filters for a band-pass peak at 10 Hz.
The CCNN here is composed of 5 layers with a kernel width of 75.
}
\label{comparison-10}
\end{figure}
\begin{figure}[H]
\centerline{\includegraphics[width=0.8\columnwidth]{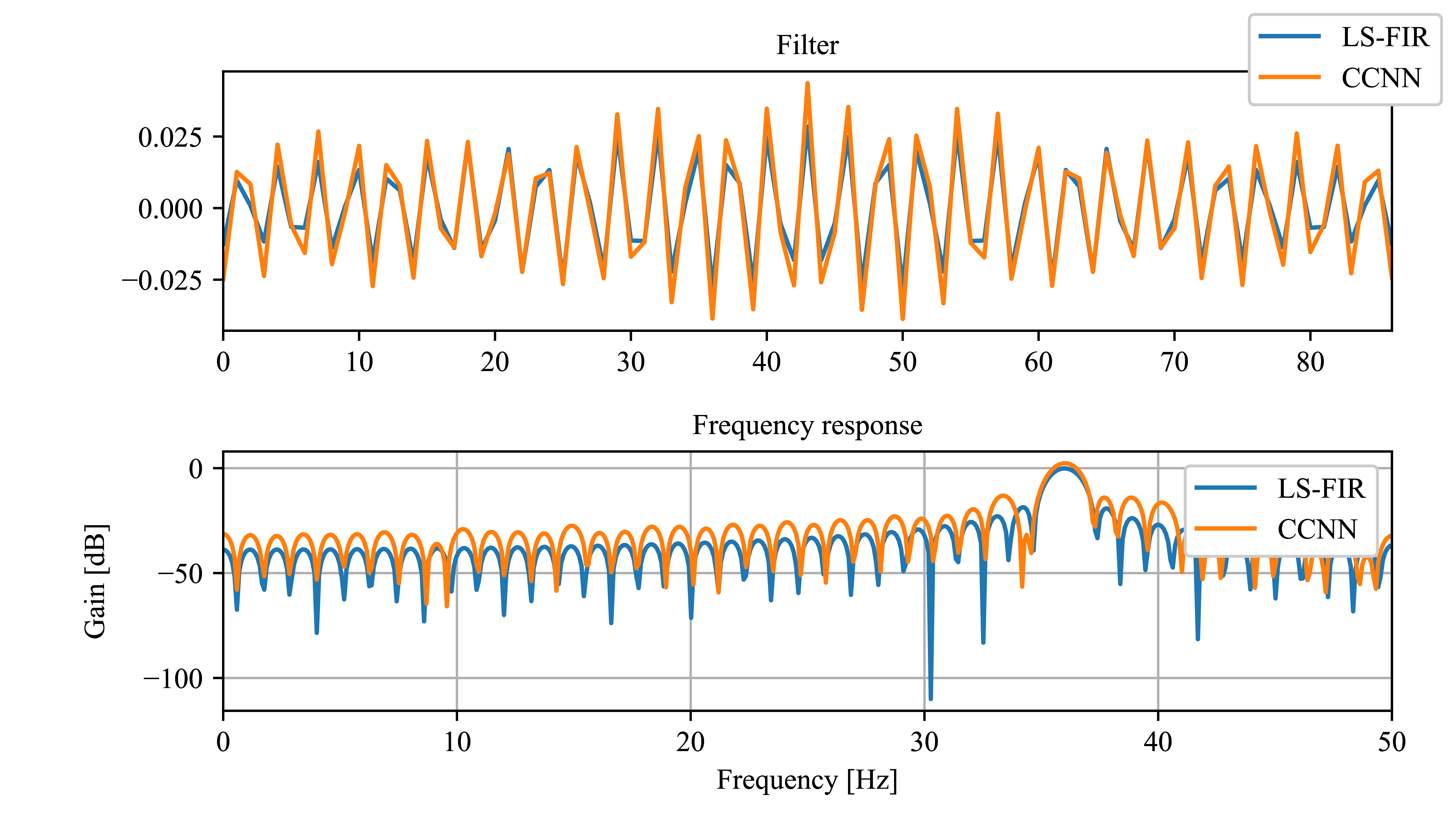}}
\caption{
A comparison of the weights and spectra of the LS-FIR and CCNN filters for a band-pass peak at 36 Hz.
The CCNN here is composed of 7 layers with a kernel width of 87.
}
\label{comparison-36}
\end{figure}
While some discrepancies are observed in terms of the gain and amplitudes of the filters, the learned filter's spectrum and weight profile remain closely aligned with those of the FIR filter obtained via the window method.

Thus, we demonstrated that a multi-layer CCNN can be combined into a single FIR filter that is optimal in the least-squares sense, whereas a single-layer CCNN fails to achieve such performance, even as the number of parameters grows exponentially.
This finding contrasts with that of \citet{Gunasekar:2018a} where full-width convolutions achieve spectral sparsity with a single layer.
Although both single-layer and deep neural networks are universal approximators \cite{Hornik:1989a} \cite{Cybenko:1989a}, they tackle the same task with fundamentally different levels of complexity.
\citet{Bianchini:2014a} demonstrated that the complexity for the single layer neural network scales exponentially, whereas it scales polynomially for a deep network.
Consistent with this theory, our experiments show that a single-layer CCNN with fixed kernel width is unable to adequately solve our baseline task (as discussed in Section~\ref{hyperparameters}).
It is possible that an increase in the number of lags could improve performance, but given our own experimental observations and in line with the works of \citet{Bianchini:2014a}, the required number of lags would need to be increased by several orders of magnitude.
In our approach, we leverage this insight by adding additional layers to reduce the effective representational complexity during training.
Post-training, we then exploit the associative property of convolution to collapse the learned layers into a single equivalent filter, thereby preserving both interpretability and spectral precision.

\subsection{System identification}
We simulate a physical multi-degree-of-freedom (MDOF) dynamical system to generate training data.
This reflects a lumped mass model, consisting of $N$ unit masses connected via linear springs of progressively increasing stiffness (from 500 to 3000~$\mathrm{N/m}$).
These stiffness values are specifically chosen to span a broad frequency range, with the fundamental mode starting at 3 Hz and extending up to the Nyquist frequency of 50 Hz.
The damping for each DOF is set in accordance with the mass/stiffness ratio such that every DOF has an effective Rayleigh damping ratio of 1\% for the first two modes.
All DOFs are subjected to the same white noise excitation.

\begin{figure}[H]
\centerline{\includegraphics[width=0.7\columnwidth]{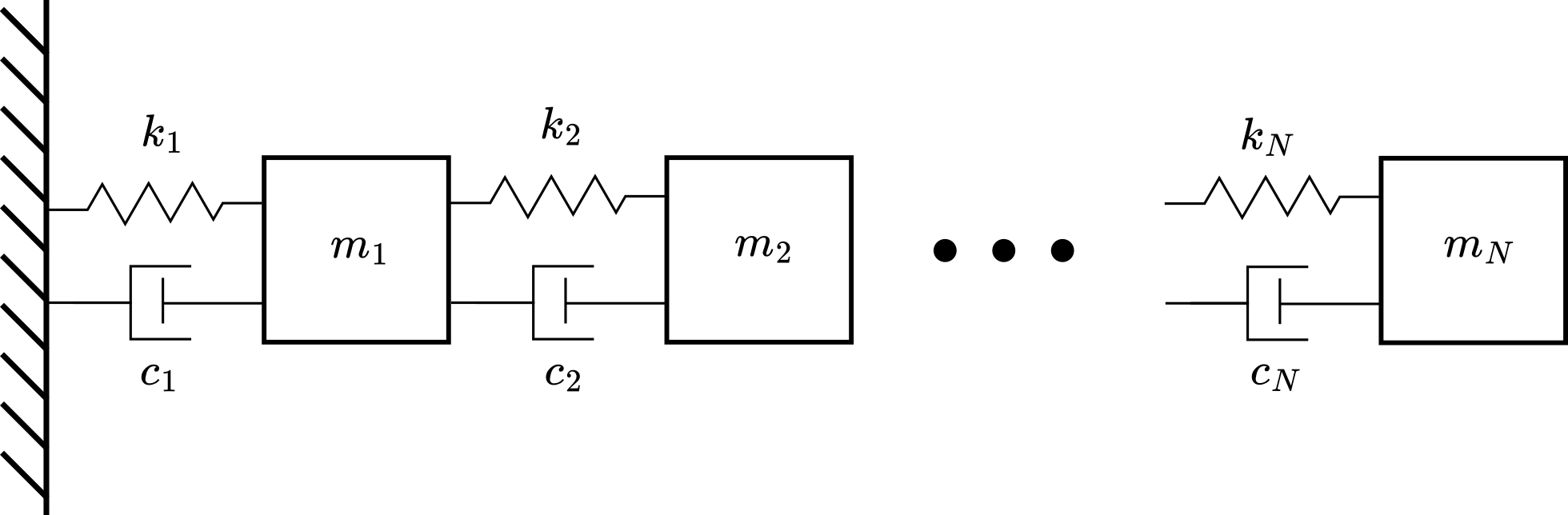}}
\caption{
The linear MDOF dynamical system used to generate the data.
The masses are set as $m_1 = m_2 = \dots = m_N = 1$.
The stiffness parameters are set such as $k_1 < k_2 < \dots < k_N \in [500, 3000]$.
The damping $c_i$ for each DOF is set such that the Rayleigh damping ratio amounts to 1\% .
}
\label{system}
\end{figure}

We simulate the system under white noise excitation of zero mean and unit variance, for a duration of 1000 seconds, at a sampling frequency of 100 Hz using a Runge-Kutta integrator. 
To construct the dataset, we extract time histories of the system’s acceleration response across all DOFs and divide them into samples of length $T = 3$ seconds.
This duration is selected to ensure that each sample contains multiple periods of the lowest natural mode, which is essential for training an autoregressive model.
Given that the lowest frequency $f_1 > 3$ Hz, a 3-second sample at $f_s = 100$ Hz captures at least 10 full cycles of the slowest mode—sufficient for effective training of our model.

\subsubsection{Regression}
\label{regression}
Next, we show that the CCNN can learn the modal characteristics of the system in an indirect manner. 
The white noise excitation is fed as input, $\mathbf{x}$, to the CCNN, with the outputs, $\mathbf{y}$, set as the accelerations of the dynamical system.
The CCNN consists of $L$ channels, which are kept separate across layers, i.e., there is no inter-channel mixing.
In addition to the CCNN, we add two expansion CNN kernels of length $p = 1$.
The first expansion kernel simply expands the forcing vector to the latent dimension of the CNN and the second projects from the latent dimension to the DOF dimension (number of acceleration channels): the full model is summarized in Figure \ref{fig:ccnn-regression}.
We project the excitation signal of length $T$ into a latent representation of dimension $L$, then pass this latent tensor through the CCNN with $M$ convolutional layers.
Finally, the output is projected back to the original DOF dimension $N$ to reconstruct the accelerations of the dynamical system.

\begin{figure}[H]
\centerline{\includegraphics[width=0.7\columnwidth]{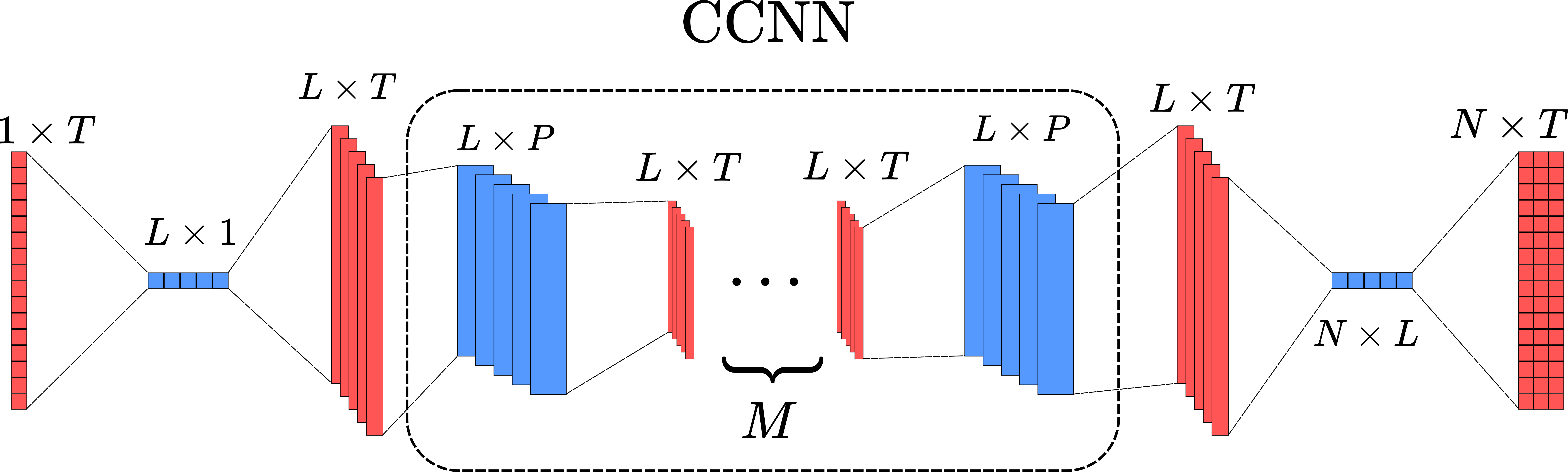}}
\caption{
The model trained for the regression problem.
The trainable weights are highlighted in blue.
The variables which are propagated forward are highlighted in red.
The batch dimension has been omitted for readability.
}
\label{fig:ccnn-regression}
\end{figure}

We train our full model on a 2DOF system with the following parameters.
The CCNN has $M = 5$ layers with $L = 2$ channels with kernels of length $p = 75$.
The model is trained for 500 epochs with a batch size of 256.
Once trained, we inspect the transfer functions of the learned CCNN filters.
As shown in Figure~\ref{fig:regression-2d}, the modes of the dynamical system manifest as distinct peaks in the frequency spectrum of each channel.
To verify the scalability of our approach, we repeat the same training procedure on a 9-DOF system (Figure~\ref{fig:regression-9d}), increasing the number of channels from $L = 2$ to $L = 3$.

\begin{figure}[h]
    \centering
    \subfloat[2 DOF\label{fig:regression-2d}]{
        \includegraphics[width=0.5\columnwidth]{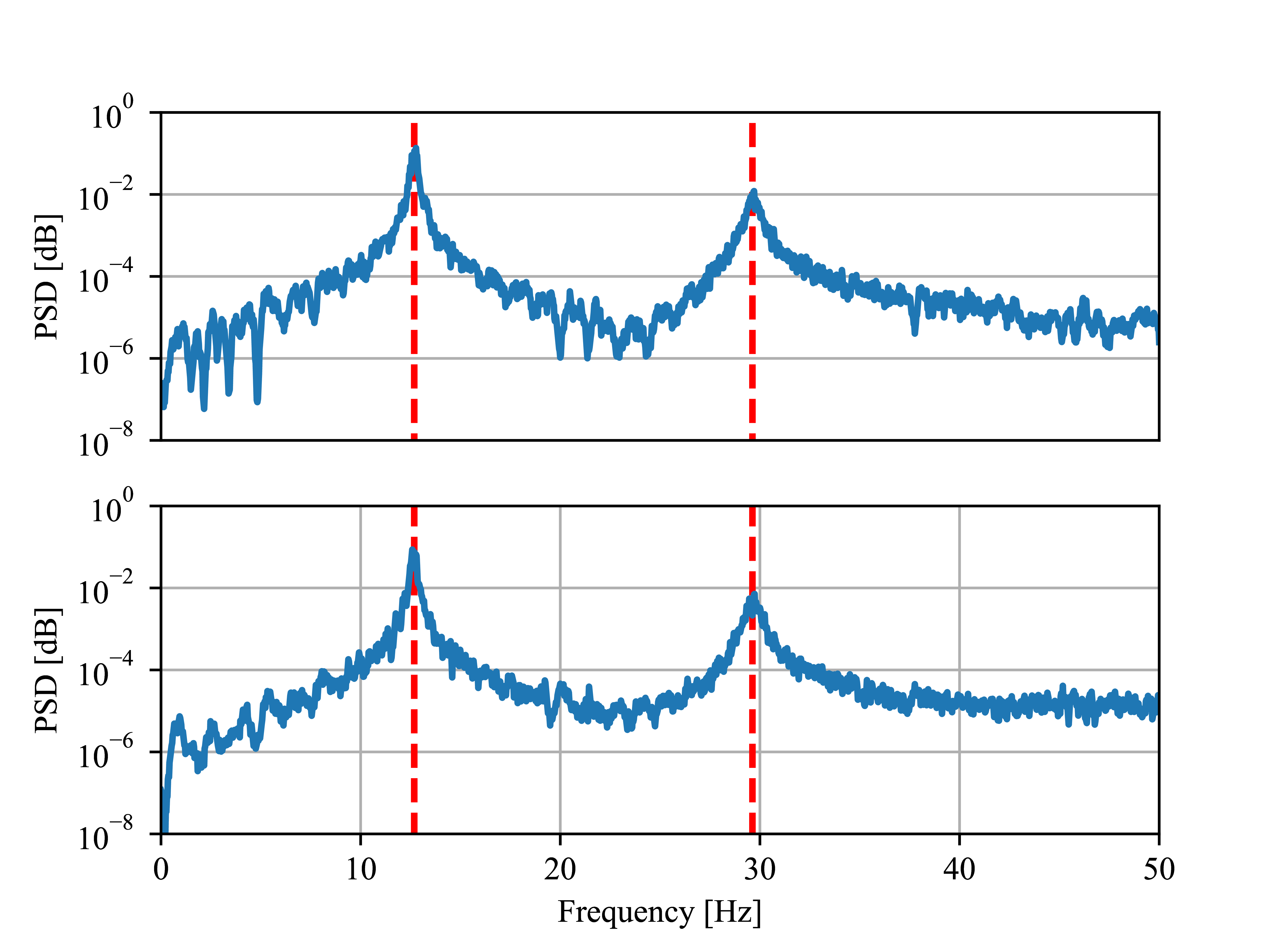}
    }
    \hfill
    \subfloat[9 DOF\label{fig:regression-9d}]{
        \includegraphics[width=0.5\columnwidth]{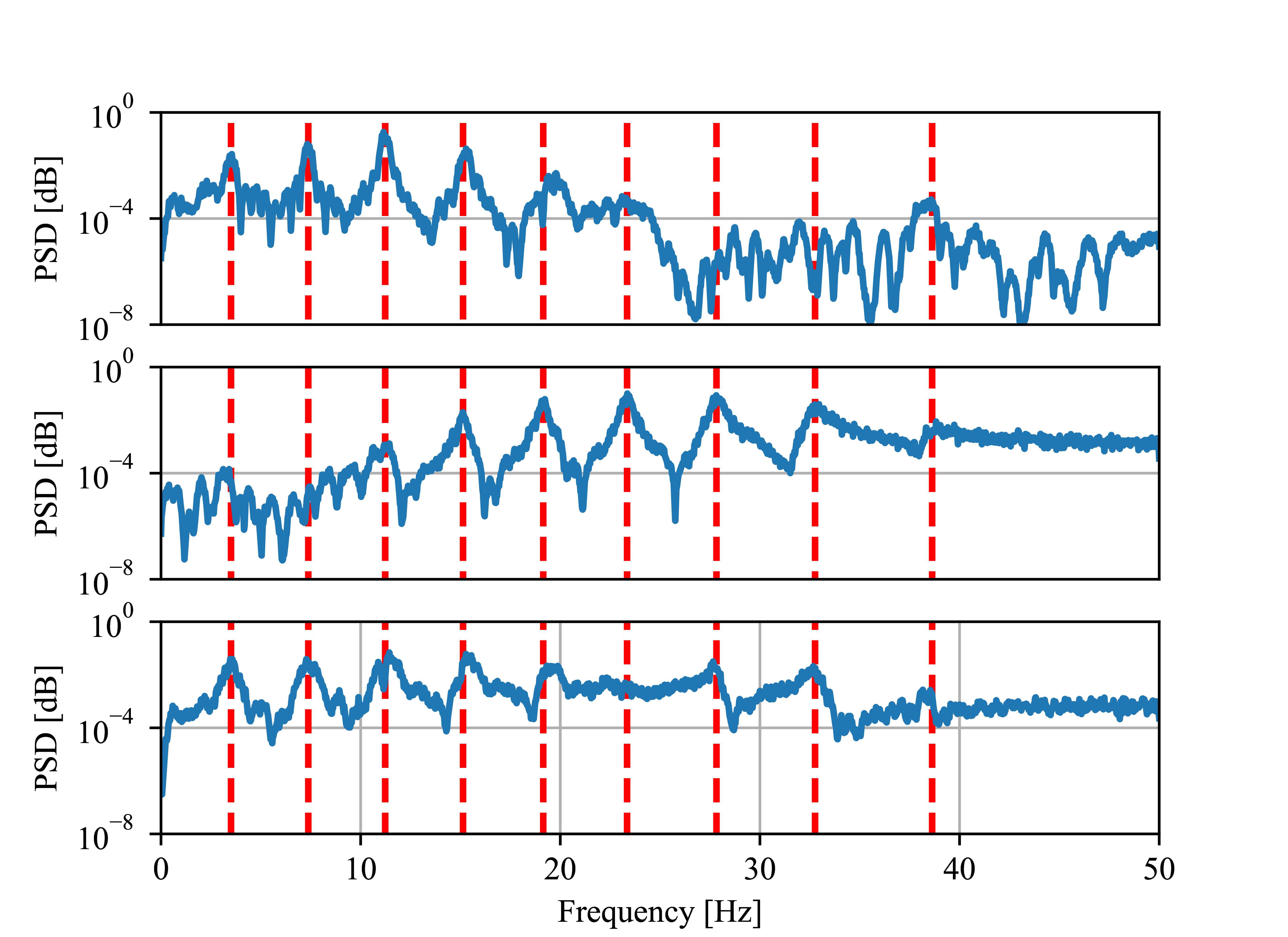}
    }
    \caption{
        Spectrum of the output for white noise excitation for the different CCNN channels for the regression task. PSD stands for Power Spectral Density.
        The true modes of the dynamic system are plotted in red dashed lines.
    }
    \label{fig:regression}
\end{figure}

These results show that the CCNN can implicitly learn the system's spectral properties.
However, it is not able to separate these natural modes, since the PSD plots of the different channels do not result as monochromatic, but instead contain multiple natural modes.
As such, the model cannot be directly compared to single-mode decomposition techniques commonly used in structural health monitoring (SHM).

\subsubsection{Unsupervised learning}
\label{unsupervised}
In the previous section, we had access to the white noise excitation data, which made it easier for the model to implicitly learn the modal properties of the system.
In this section, we examine whether the CCNN can learn these properties in an unsupervised setting.
To this end, we train a Variational Autoencoder \cite{Kingma:2013a} that learns a latent representation of the acceleration, using a  Long Short-Term Memory (LSTM) network as an encoder and a CCNN as a decoder.
The VAE latent space prior is assumed Gaussian, which encourages a white latent space.
Under this premise, we expect the decoder CCNN to act similarly to the regressor model trained in the previous section.
An LSTM is used for encoding specifically to avoid introducing any spectral bias into the latent representation.
The full model is summarized in Figure \ref{ccnn-unsupervised}.
We train our model on the same acceleration data $\mathbf{y}$, used previously, but without access to the corresponding excitation signals $\mathbf{x}$.
The LSTM extracts features from the acceleration time histories and maps them to the mean $\mu$ and standard deviation $\sigma$ of the VAE’s latent distribution.
A sample drawn from this distribution is then passed through the CCNN decoder consisting of $M$ layers.
The output is finally projected to the number of DOFs, $N$, to reconstruct the system’s acceleration response.
The model is trained using the same hyperparameters as in the regression task, and the LSTM consists of 2 layers with 20\% dropout.

\begin{figure}[H]
\centerline{\includegraphics[width=0.85\columnwidth]{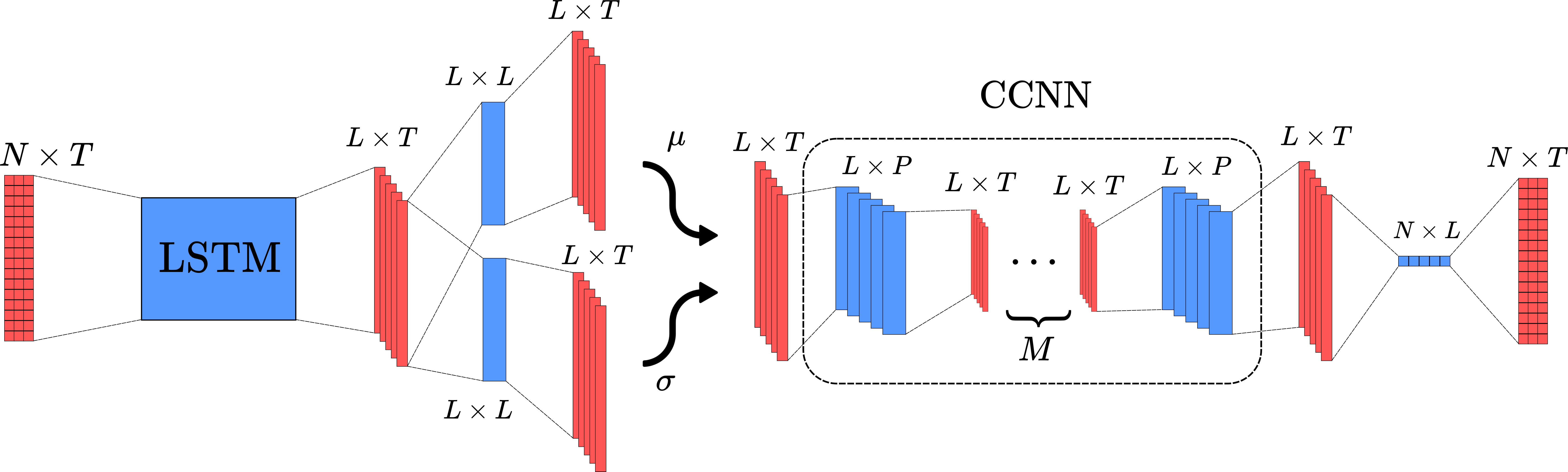}}
\caption{
The model trained for the unsupervised learning problem.
The trainable weights are highlighted in blue.
The variables which are propagated forward are highlighted in red.
The batch dimension has been omitted for readability.
}
\label{ccnn-unsupervised}
\end{figure}

The spectrum of the CCNN is shown in Figure \ref{fig:unsupervised-2d}.
We observe that the natural modes of the dynamical system appear in the spectrum of the trained filters, in absence of any information on the excitation.
Additionally, we note that the natural modes have been separated, since there exists a main dominant peak in the resulting channel spectra.
Similarly to the previous section, we verify the scalability of our solution by increasing the dimension of the problem through an increase in the employed DOFs, $N$, from 2 to 9 DOFs.
The spectrum of the resulting CCNN is shown in Figure \ref{fig:unsupervised-9d}.

\begin{figure}[h]
    \centering
    \subfloat[2 DOF linear system\label{fig:unsupervised-2d}]{%
        \includegraphics[width=0.5\columnwidth]{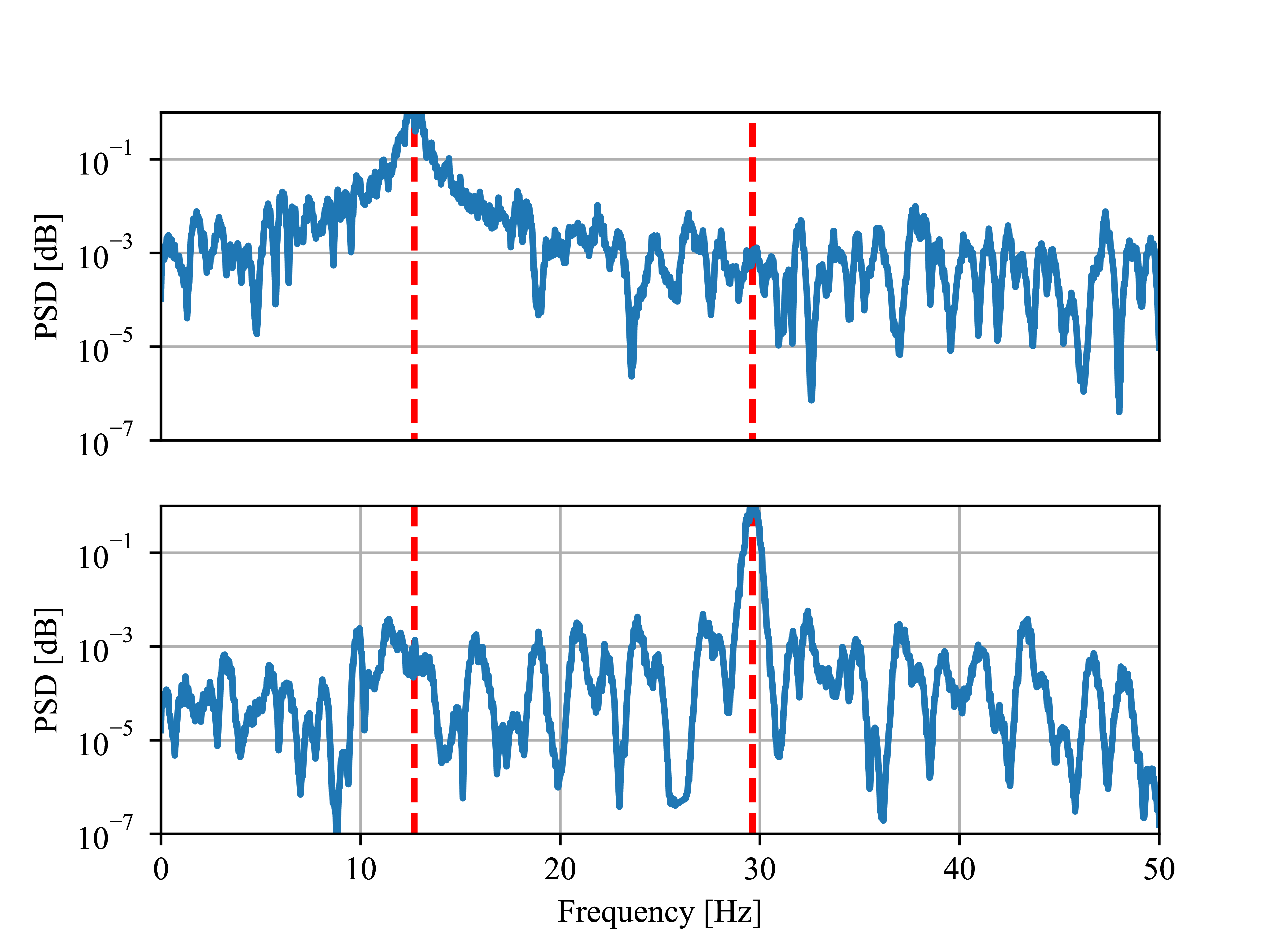}%
    }
    \hfill
    \subfloat[9 DOF linear system\label{fig:unsupervised-9d}]{%
        \includegraphics[width=0.5\columnwidth]{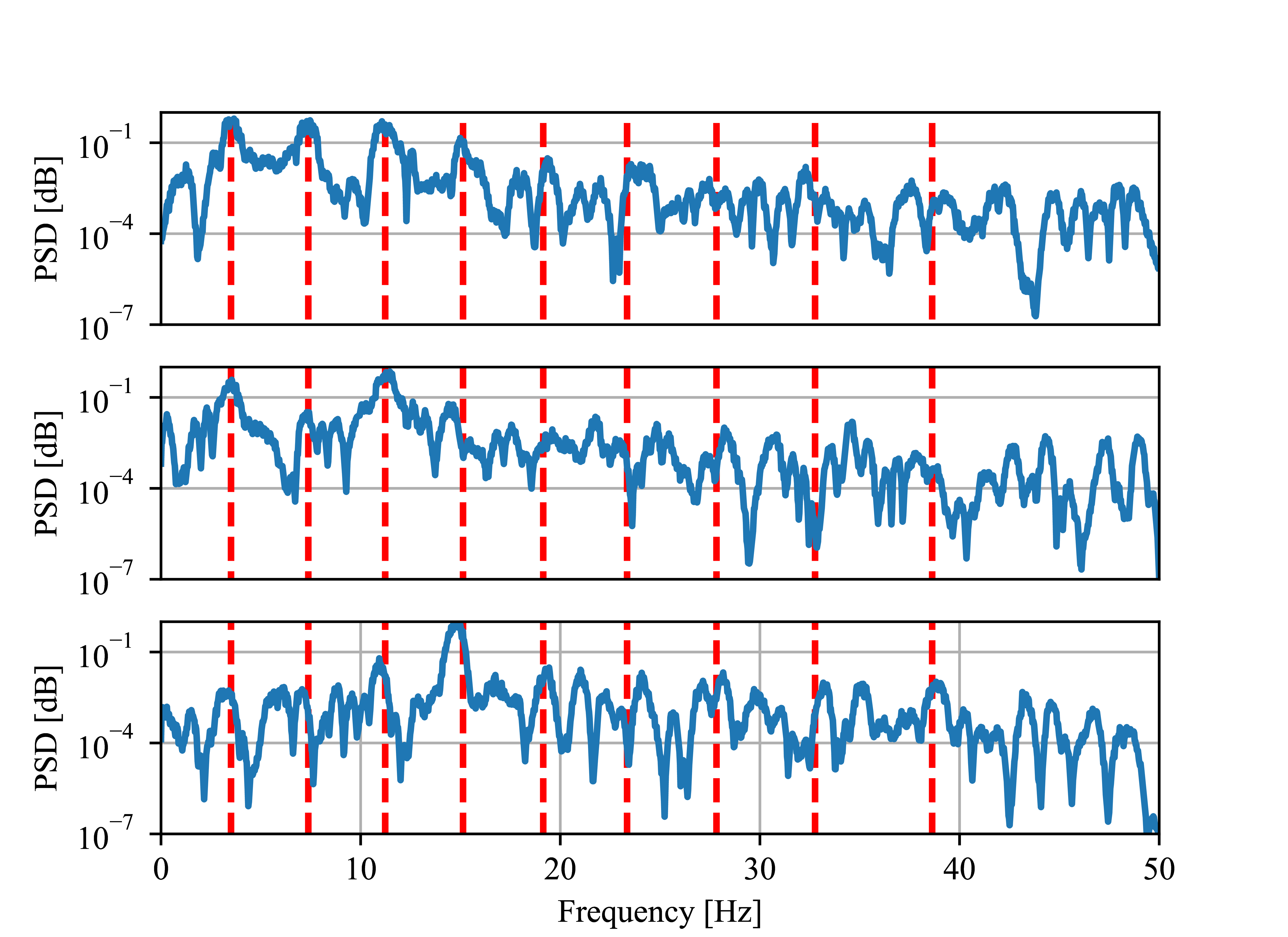}%
    }
    \caption{
        Spectrum of the output for white noise excitation for the different CCNN channels for the unsupervised learning task.
        PSD stands for Power Spectral Density.
        The true modes of the dynamic system are plotted in red dashed lines.
    }
    \label{fig:unsupervised}
\end{figure}

In the 9 DOF case, only the first 4 natural modes are reliably recovered.
This is not unexpected when considering the spectral energy distribution of the system itself, shown in Figure~\ref{fig:original-9d}.
\begin{figure}[H]
\centering
\centerline{\includegraphics[width=\columnwidth]{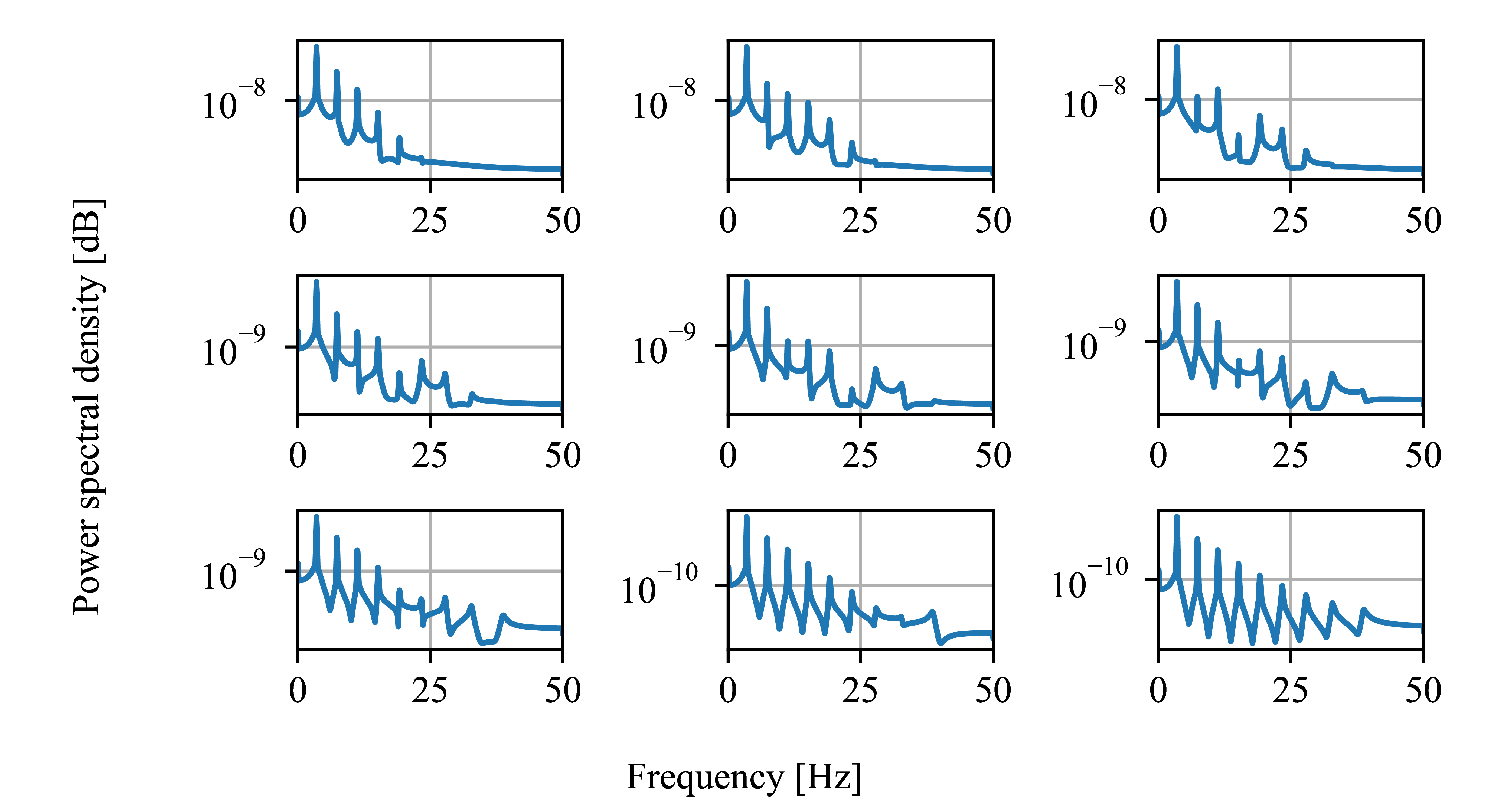}}
\caption{
Spectrum of the impulse response for each DOF of a 9 DOF dynamical system.
}
\label{fig:original-9d}
\end{figure}
It is evident that several natural modes are already suppressed in the raw system dynamics.
Consequently, recovering these modes through an unsupervised learning scheme, without access to excitation data, proves to be particularly challenging.

\subsection{Z24 bechmark}

In this section, we validate our approach on real-world data.
We selected the Z24 bridge benchmark \cite{Maeck2003} as a case study to validate our unsupervised learning scheme on structural vibrational data.
Built in 1963, the Z24 bridge (see Figure \ref{fig:z24}) was located in Solothurn, Switzerland.
It was demolished in 1998 to allow for the widening of the neighboring railway bridge.
Prior to its demolition, the bridge was subjected to a forced operational test. 
An array of 291 accelerometers was deployed to collect vibrational data along the bridge as a 1 kN shaker generated vibrations with a flat spectrum between 3 and 30 Hz.
No data was provided regarding the shaker input.
As a reference, we consider the expected natural frequencies reported by \citet{Maeck2003}:
[3.87, 4.82, 9.77, 10.5, 12.4, 13.2, 17.2, 19.3, 19.8, 26.6] Hz.
The data was originally sampled at 100 Hz.
For our analysis, we downsampled to 60 Hz using anti-aliasing filters, focusing on frequencies below 30 Hz.
Although the dataset contains records under various simulated damage conditions, we restrict our analysis to the baseline (undamaged) bridge response.
\begin{figure}[ht]
\centering
\centerline{\includegraphics[width=0.7\columnwidth]{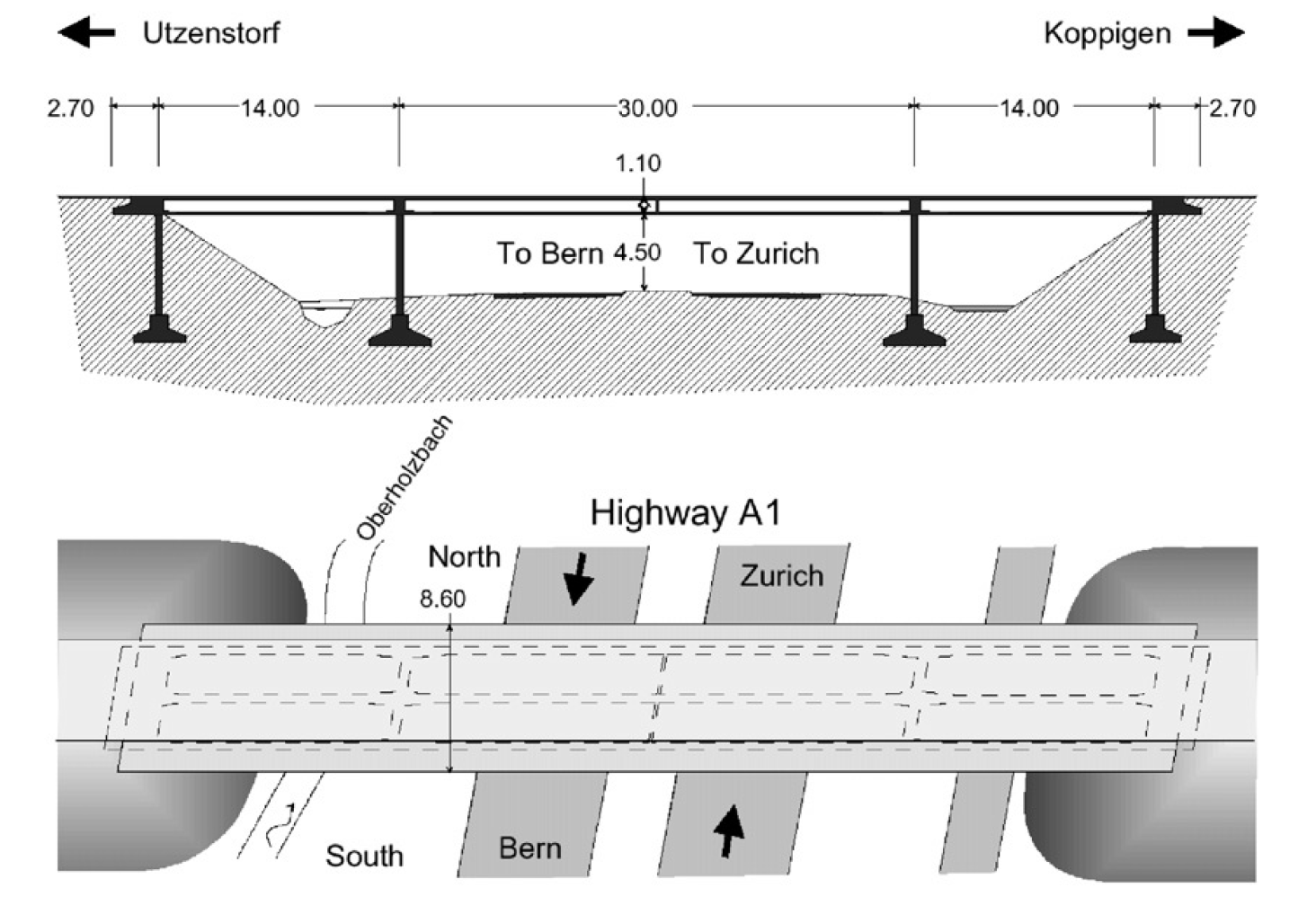}}
\caption{
Diagram of the Z24 bridge provided by \citet{Maeck2003}.
}
\label{fig:z24}
\end{figure}

Before applying our CCNN-based method, we first assess which modes can be reliably recovered using Frequency Domain Decomposition (FDD) \cite{FDD2000}.
The FDD method begins by computing the cross power spectral density (CPSD) matrix of the multichannel acceleration signals.
The CPSD between two signals, $x(t)$ and $y(t)$, is defined as the Fourier transform of their cross-correlation:

\begin{equation}
    S_{xy}(f) = \int_{-\infty}^{\infty} \mathbb{E}[x(t) y^*(t - \tau)] e^{-2 \pi i f \tau} d \tau
\end{equation}

The full CPSD matrix $\mathbf{S}(f)$ is a Hermitian matrix composed of all pairwise spectral densities:

\begin{equation}
    \mathbf{S}(f) = \begin{bmatrix}
S_{x_1 x_1} (f) & S_{x_1 x_2} (f) & \dots & S_{x_1 x_N} (f) \\
S_{x_2 x_1} (f) & S_{x_2 x_2} (f) & \dots & S_{x_2 x_N} (f) \\
\vdots & \vdots & \ddots & \vdots \\
S_{x_N x_1} (f) & S_{x_N x_2} (f) & \dots & S_{x_N x_N} (f) 
\end{bmatrix}
\end{equation}

Although CPSD is theoretically defined via continuous-time Fourier transforms over infinitely long stationary processes, in practice we apply Welch’s method, which is more suitable for finite-length, ergodic signals.
We use a window length of $2048 \Delta t$ with 50\% overlap, resulting in 1025 frequency bins and a spectral resolution of approximately 0.0293 Hz.
We compute the CPSD over the entire Z24 dataset samples that are recorded while the Z24 bridge structure is still considered healthy.
This dataset amounts to 1730 seconds of data, approximately 28.83 minutes, which is long enough for Operational Modal Analysis (OMA) to be carried out on ambient vibration data.
We then compute the singular value decomposition for matrix $\mathbf{S}$ of each window and extract the top 10 singular values.
We can plots these singular values over the frequency spectrum, as per Figure \ref{fig:z24-psd}.
Additionally, we overlay the true natural modes of the bridge given by \citet{Maeck2003}.
\begin{figure}[ht]
\centering
\centerline{\includegraphics[width=0.9\columnwidth]{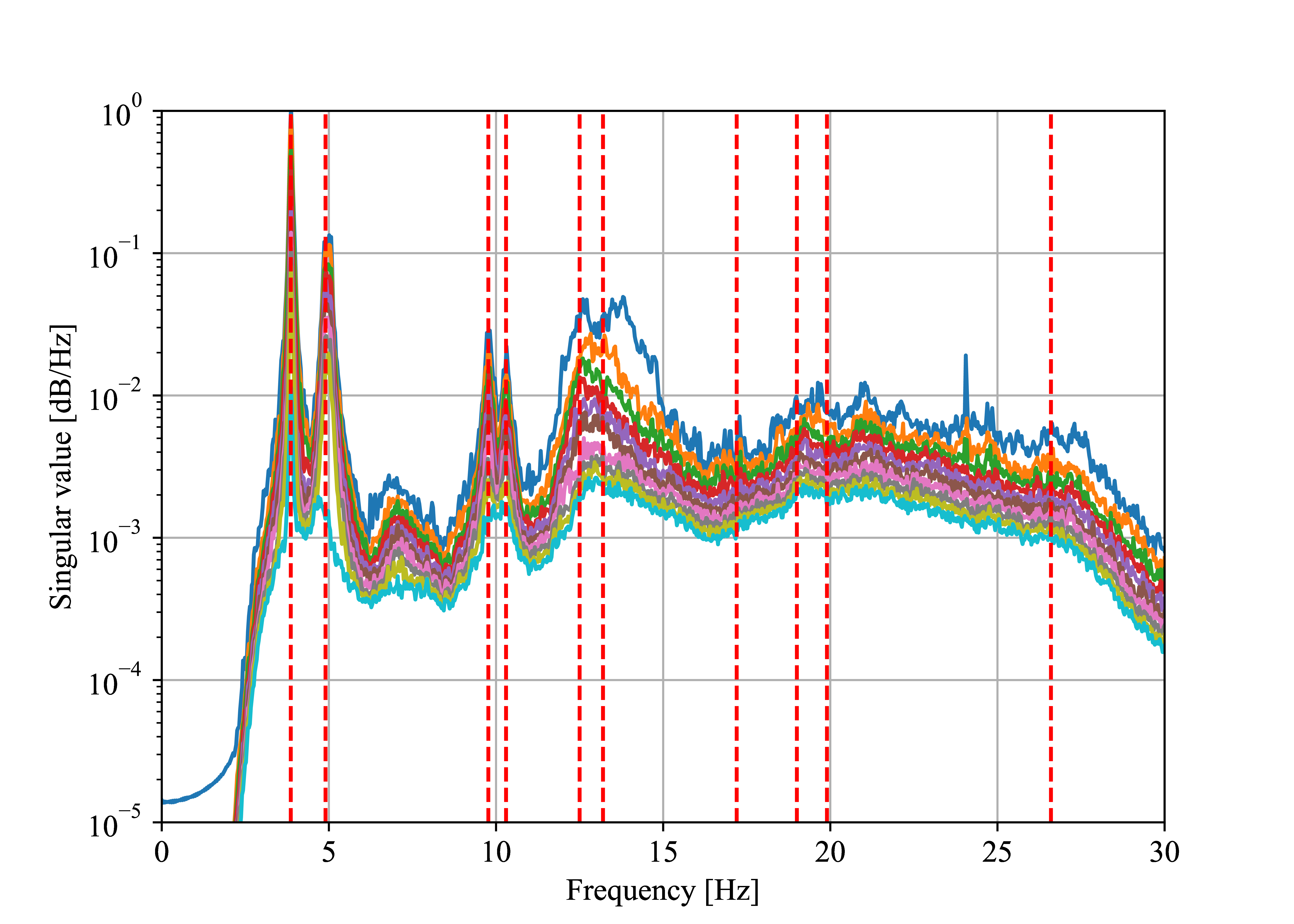}}
\caption{
Singular values of the CPSD of the vibrational signals of the Z24 bridge dataset.
The true natural modes of the bridge are plotted in red dashed lines.
}
\label{fig:z24-psd}
\end{figure}
We observe that many of the natural modes are not prominently visible in the recorded dynamics, making recovery challenging.
While the lower modes (below ~12 Hz) appear clearly, higher modes become increasingly difficult to identify.
Additionally, the lower peaks are closely spaced, which poses a challenge for separation using our proposed CCNN.
The CCNN is trained in an unsupervised scheme in the same manner as in the previous section.
The structure of the LSTM and the VAE remain unchanged.
The CCNN is this case is designed with 8 filters with 8 layers with a kernel length of 201, using tanh activations.
The time-history records of the Z24 benchmark are divided into samples of 3000 time steps (50 seconds).
\begin{figure}[ht]
\centering
\centerline{\includegraphics[width=\columnwidth]{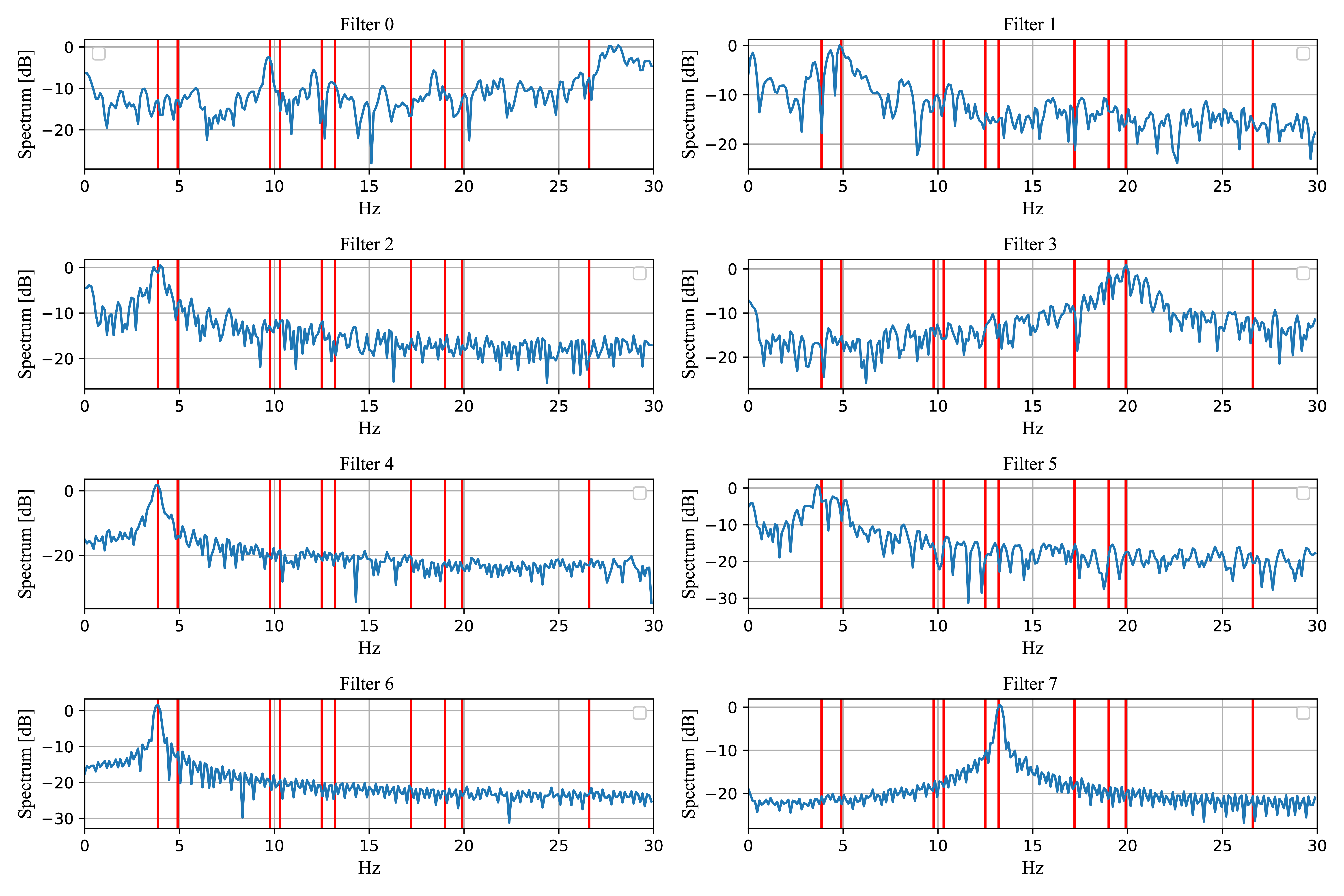}}
\caption{
Filters learned by the CCNN from the Z24 data.
The true natural modes of the bridge are plotted in red.
}
\label{fig:z24-filters}
\end{figure}
To compare our approach with FDD, we sum the spectra of all CCNN channels and overlay them with the first singular value of the CPSD.
To reduce ripple artifacts, we apply a Savitzky–Golay filter \cite{Savitzky1964} with a window of 51 and polynomial order 3.
The comparison is given in Figure \ref{fig:z24-comparison}.
While both methods recover the natural modes of the Z24 bridge, there are a few notable differences.
The CCNN produces a sparser spectrum, making the modal peaks more prominent than in the FDD result.
Importantly, the CCNN achieves this using only 50-second samples, enabling localized detection of modal changes during deployment—unlike FDD, which requires long durations (e.g., the full 28 minutes) to achieve reliable results.
On the other hand, the CCNN spectra show slight deviations in peak frequency when compared to the FDD baseline, reflecting a trade-off between interpretability and precision.

\begin{figure}[ht]
\centering
\centerline{\includegraphics[width=0.9\columnwidth]{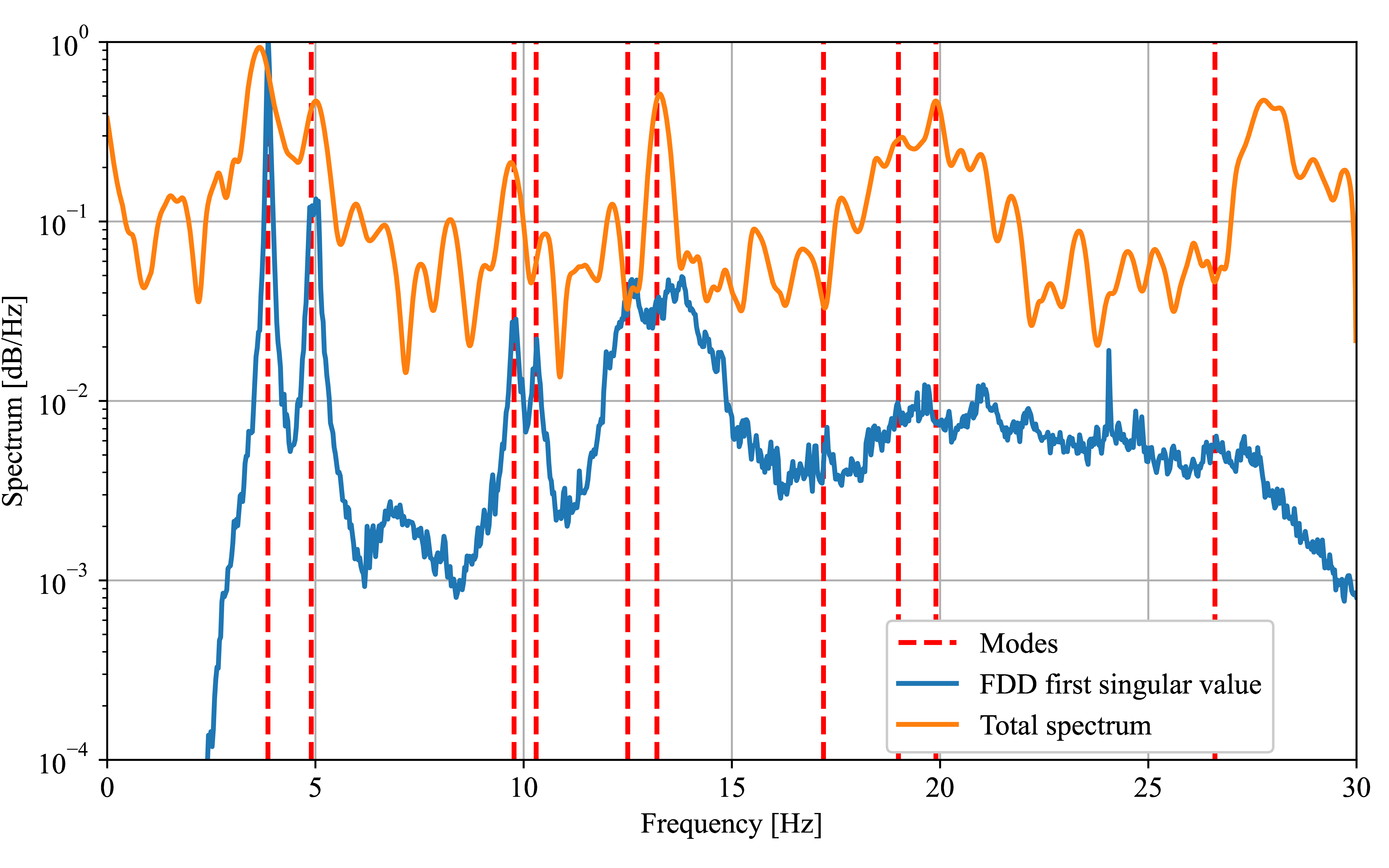}}
\caption{
A comparison with the FDD and the total CCNN spectrum  (sum of all channel spectra)that are obtained by analyzing the Z24 dataset.
}
\label{fig:z24-comparison}
\end{figure}


\section{Conclusion}
In this work, we explored the behavior of Causal Convolutional Neural Networks (CCNNs) when applied to the analysis of dynamic systems.
Through a range of illustrative tasks, including spectrum learning, regression, and unsupervised identification, we demonstrated that CCNNs trained via gradient-based optimization converge toward behavior closely resembling that of Least-Squares Finite Impulse Response (LS-FIR) filters.
This connection becomes particularly evident when using long convolutional kernels, which allow the networks to capture narrow-band spectral features commonly observed in structural vibration data.

A key insight from our study is the emergence of a spectral bias in CCNNs.
This bias arises explicitly when the network is trained directly on band-limited signals, and implicitly when the model is tasked with reconstructing or encoding the dynamic response of physical systems—without any frequency supervision.
In both cases, the network learns to concentrate its filtering capacity around dominant modal frequencies, which are fundamental to characterizing system dynamics.
We also show that using multiple quasi-linear convolutional layers not only improves convergence but also allows the entire network to be collapsed post-training into a single interpretable filter.
This structural simplicity does not compromise performance and, in fact, enables a direct connection to classical signal processing concepts, such as linear-phase filtering and modal decomposition.
The practical value of this framework is highlighted through applications to both simulated MDOF systems and real vibrational data from the Z24 bridge benchmark.
In particular, the CCNN demonstrated the ability to identify key modal components using short-duration signal segments—offering a lightweight and interpretable alternative to traditional spectral analysis techniques such as Frequency Domain Decomposition (FDD).
However, it should be noted that our goal is not to propose this method as a replacement for established identification techniques, but rather to provide insight into how neural networks, particularly CNNs, process and internalize dynamic signals, enabling better-informed use of such models in the context of physical systems.
By combining spectral interpretability with the flexibility of deep learning, CCNNs present a promising tool for structural health monitoring and related tasks involving time-series data from physical systems.

\section*{Acknowledgments}
The research was conducted as part of the Future Resilient Systems (FRS) program at the Singapore-ETH Centre, which was established collaboratively between ETH Zurich and the National Research Foundation Singapore. 
This research is supported by the National Research Foundation, Prime Minister’s Office, Singapore under its Campus for Research Excellence and Technological Enterprise (CREATE) programme.

{\appendix[Least-square FIR filter]
For our baseline comparison filter, we choose a FIR filter that has been optimized to target a linearized bandwidth using a least-square optimization scheme.
The following is based on the derivations from \citet{Selesnick2005}.
With $A(\omega)$ the response of the filter, $D(\omega)$ the ideal response and $\mathcal{W}(\omega)$ the loss weighing function, we seek to optimize the weighted integral squared error:

\begin{equation}
    \mathcal{E}_2 = \int_{0}^{\pi} \mathcal{W}(\omega) (A(\omega) - D(\omega))^2 d \omega
\end{equation}

Where we have, for $\mathcal{P} = \frac{p - 1}{2}$:

\begin{equation}
    A(\omega) = \sum_{n=0}^\mathcal{P} a(n) \cos (n \omega)
\end{equation}

We would like to minimize $\mathcal{E}_2$ with respect to $a (n)$, i.e. we want:

\begin{equation}
    \frac{d \mathcal{E}_2}{d a (k)} = 0, 0 \leq k \leq \mathcal{P}
\end{equation}

which yields the following:

\begin{multline}
    \frac{d \mathcal{E}_2}{d a (k)} = \int_{0}^{\pi} \frac{d}{d a(k)} \mathcal{W}(\omega) (A(\omega) - D(\omega))^2 d \omega \\
    = 2 \int_{0}^{\pi} \mathcal{W}(\omega) (A(\omega) - D(\omega)) \frac{d A}{d a(k)} d \omega \\
    = 2 \int_{0}^{\pi} \mathcal{W}(\omega) (A(\omega) - D(\omega)) \cos (k \omega) d \omega
\end{multline}

Therefore, $\frac{d \mathcal{E}_2}{d a (k)} = 0$ is equivalent to:

\begin{equation}
    \int_{0}^{\pi} \mathcal{W}(\omega) A(\omega) \cos (k \omega) d \omega = \int_{0}^{\pi} \mathcal{W}(\omega) D(\omega) \cos (k \omega) d \omega
\end{equation}

or alternatively:

\begin{equation}
    \sum_{n=0}^{\mathcal{P}} a(n) \int_{0}^{\pi} \mathcal{W}(\omega) \cos (n \omega) \cos (k \omega) d \omega = \int_{0}^{\pi} \mathcal{W}(\omega) D(\omega) \cos (k \omega) d \omega
\end{equation}

If we define $Q(k, n)$ and $b(k)$ such that:

\begin{equation}
    Q(k, n) = \frac{1}{\pi} \int_{0}^{\pi} \mathcal{W}(\omega) \cos (n \omega) \cos (k \omega) d \omega
\end{equation}

and

\begin{equation}
    b(k) = \frac{1}{\pi} \int_{0}^{\pi} \mathcal{W}(\omega) D(\omega) \cos (k \omega) d \omega
\end{equation}

then the optimization of the filter can be written as a system of linear equations, such that:

\begin{equation}
    \begin{bmatrix}
        Q(0, 0) & Q(0, 1) & ... & Q(0, \mathcal{P}) \\
        Q(1, 0) & Q(1, 1) & ... & Q(1, \mathcal{P}) \\
        \vdots & &  & \vdots \\
        Q(\mathcal{P}, 0) & Q(\mathcal{P}, 1) & ... & Q(\mathcal{P}, \mathcal{P}) \\
        \end{bmatrix}
    \begin{bmatrix}
        a(0) \\
        a(1)  \\
        \vdots \\
        a(\mathcal{P}) \\
    \end{bmatrix} =  
    \begin{bmatrix}
        b(0) \\
        b(1)  \\
        \vdots \\
        b(\mathcal{P}) \\
    \end{bmatrix} 
\end{equation}

Therefore, the FIR filter can be obtained by solving the least-square problem $a = Q^{-1} b$.

The matrix $Q$ can be further simplified.
Using the following equality:

\begin{equation}
    \cos (n \omega) \cos (k \omega) = \frac{1}{2} \cos ((k - n) \omega) + \frac{1}{2} \cos ((k + n) \omega)    
\end{equation}

$Q(k, n)$ can be rewritten as follows:

\begin{multline}
    Q(k, n) = \frac{1}{\pi} \int_{0}^{\pi} \mathcal{W}(\omega) \cos (n \omega) \cos (k \omega) d \omega \\ =  
    \frac{1}{2 \pi} \int_{0}^{\pi} \mathcal{W}(\omega) \cos ((k-n) \omega) d \omega + 
    \frac{1}{2 \pi} \int_{0}^{\pi} \mathcal{W}(\omega) \cos ((k-n) \omega) d \omega \\ =
    \frac{1}{2} Q_1 (k, n) + \frac{1}{2} Q_2 (k, n)
\end{multline}

If we define $q$ such that:

\begin{equation}
    q(n) = \frac{1}{\pi} \int_{0}^{\pi} \mathcal{W}(\omega) \cos (n \omega) d \omega
\end{equation}

Then we can write $Q_1 (k, n) = q(k - n)$ and $Q_2 (k, n) = q(k + n)$. 
Therefore, matrices $Q_1$ and $Q_2$ can be rewritten as:

\begin{equation}
    Q_1 = \begin{bmatrix}
        q(0) & q(1) & ... & q(\mathcal{P}) \\
        q(1) & q(0) & ... & q(\mathcal{P}-1) \\
        \vdots & &  & \vdots \\
        q(\mathcal{P}) & q(\mathcal{P}-1) & ... & q(0) \\
        \end{bmatrix}
\end{equation}

\begin{equation}
    Q_2 = \begin{bmatrix}
        q(0) & q(1) & ... & q(\mathcal{P}) \\
        q(1) & q(2) & ... & q(\mathcal{P}+1) \\
        \vdots & &  & \vdots \\
        q(\mathcal{P}) & q(\mathcal{P}+1) & ... & q(2\mathcal{P}) \\
        \end{bmatrix}
\end{equation}

Which allows for more efficient solving \cite{Heining2011}.

So far we have assumed an arbitrary $\mathcal{W}(\omega)$ and $D(\omega)$. 
To find a closed form solution, we introduce the following assumptions:

\begin{itemize}
    \item We solve the least-square optimization on sub-bands of the spectrum.
    \item For each band, $\mathcal{W}(\omega) = \mathcal{W}$ is constant.
    \item For each band, $D(\omega) = m \frac{\omega}{\pi} + c$ is linearized.
\end{itemize}

We first find a closed form for $q$. 
With a constant $\mathcal{W}$, $q$ can be calculated as:

\begin{multline}
    q(n) = \frac{1}{\pi} \int_{0}^{\pi} \mathcal{W}(\omega) \cos (n \omega) d \omega = \mathcal{W} \int_{0}^{1} \cos (n \pi f) df = \\ \mathcal{W} f \frac{\sin (n \pi f)}{n \pi f} |_0^1
\end{multline}

Next we find a closed form for $b$.
\begin{multline}
    b(k) = \frac{1}{\pi} \int_{0}^{\pi} \mathcal{W}(\omega) D(\omega) \cos (k \omega) d \omega = \\ \mathcal{W} \frac{1}{\pi} \int_{0}^{\pi} (m \frac{\omega}{\pi} + c) \cos (k \omega) d \omega \\ = \mathcal{W} \int_{0}^{1} (m f + c) \cos (k \pi f) df \\ = \mathcal{W} (f (m f + c) \frac{\sin (k \pi f)}{k \pi f} + m f^2 \frac{\cos (k \pi f)}{(k \pi f)^2})|_0^1
\end{multline}

This allows us to calculate $Q_1$, $Q_2$ and $b$, allowing us to solve for $a$.
Once we have the values of $a$, we can force our final filter $W$ to be symmetric by setting the coefficients as follows:

\begin{equation}
    W = [a_\mathcal{P}, a_{\mathcal{P}-1}, ..., a_1, 2 a_0, a_1, ..., a_{\mathcal{P}-1}, a_\mathcal{P}]
\end{equation}

}
\appendix[Effects of parameters and activations]{
\label{hyperparameters}

In this section, We evaluate the sensitivity to hyperparameters of the CCNN on the task in section \ref{target-spectrum}.
The parameter sweep for the number of layers $M$ and the kernel size $p$ is given in figures \ref{fig:layer-sweep} and \ref{fig:kernel-sweep}.

\begin{figure}[h]
\centering
\includegraphics[width=\columnwidth]{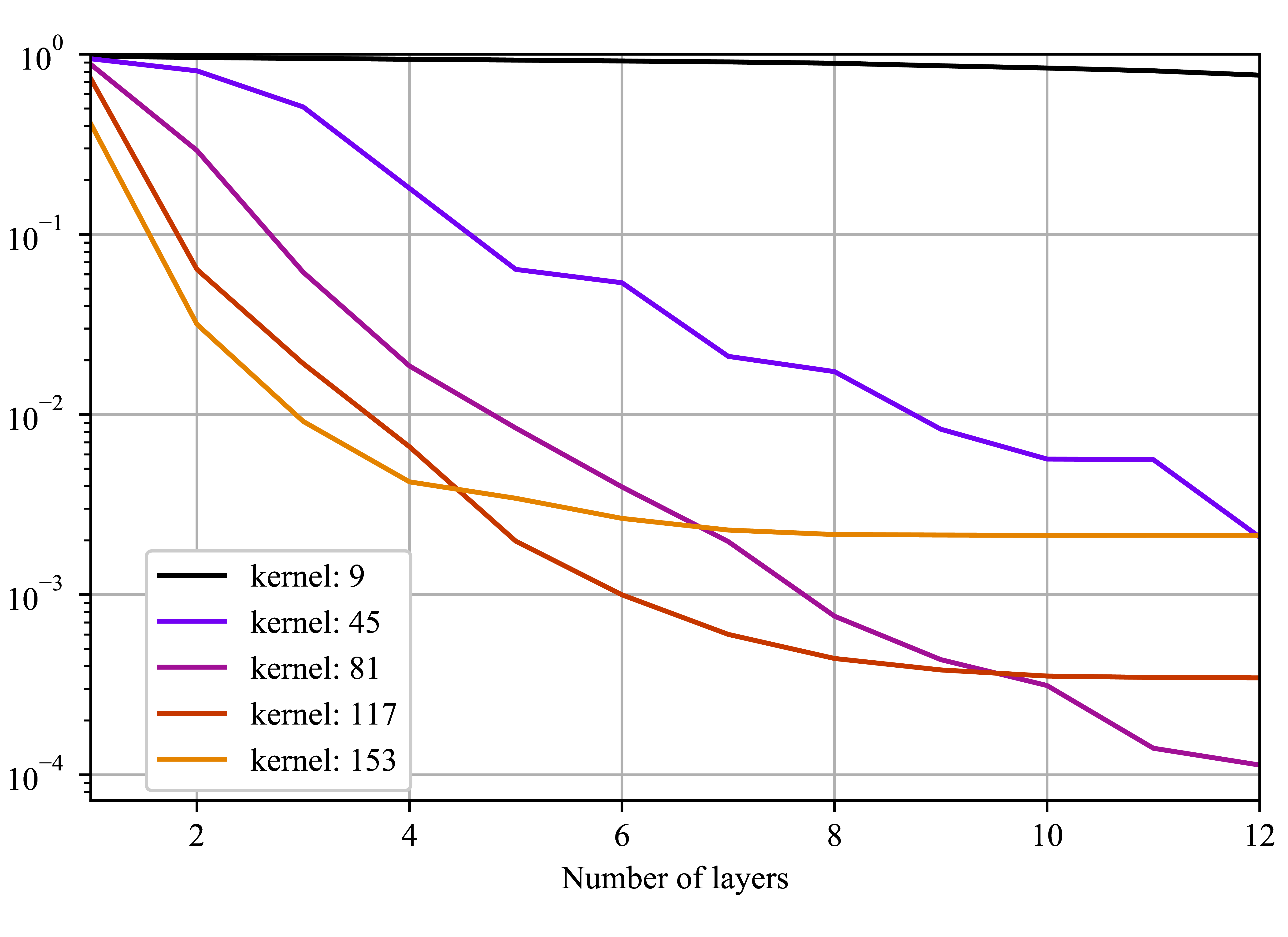}
\caption{
Loss of the CCNN the spectrum learning task for different kernel sizes as a function of the number of layers.
}
\label{fig:layer-sweep}
\end{figure}

Figure \ref{fig:kernel-sweep} confirms our on proposition that longer kernels are necessary to learn spectral biased data.
Furthermore, models with more parameters plateau after a certain number of additional layers, as they are faced with the vanishing gradient problem which diminishes the contribution of additional layers.

\begin{figure}[h]
\centering
\includegraphics[width=\columnwidth]{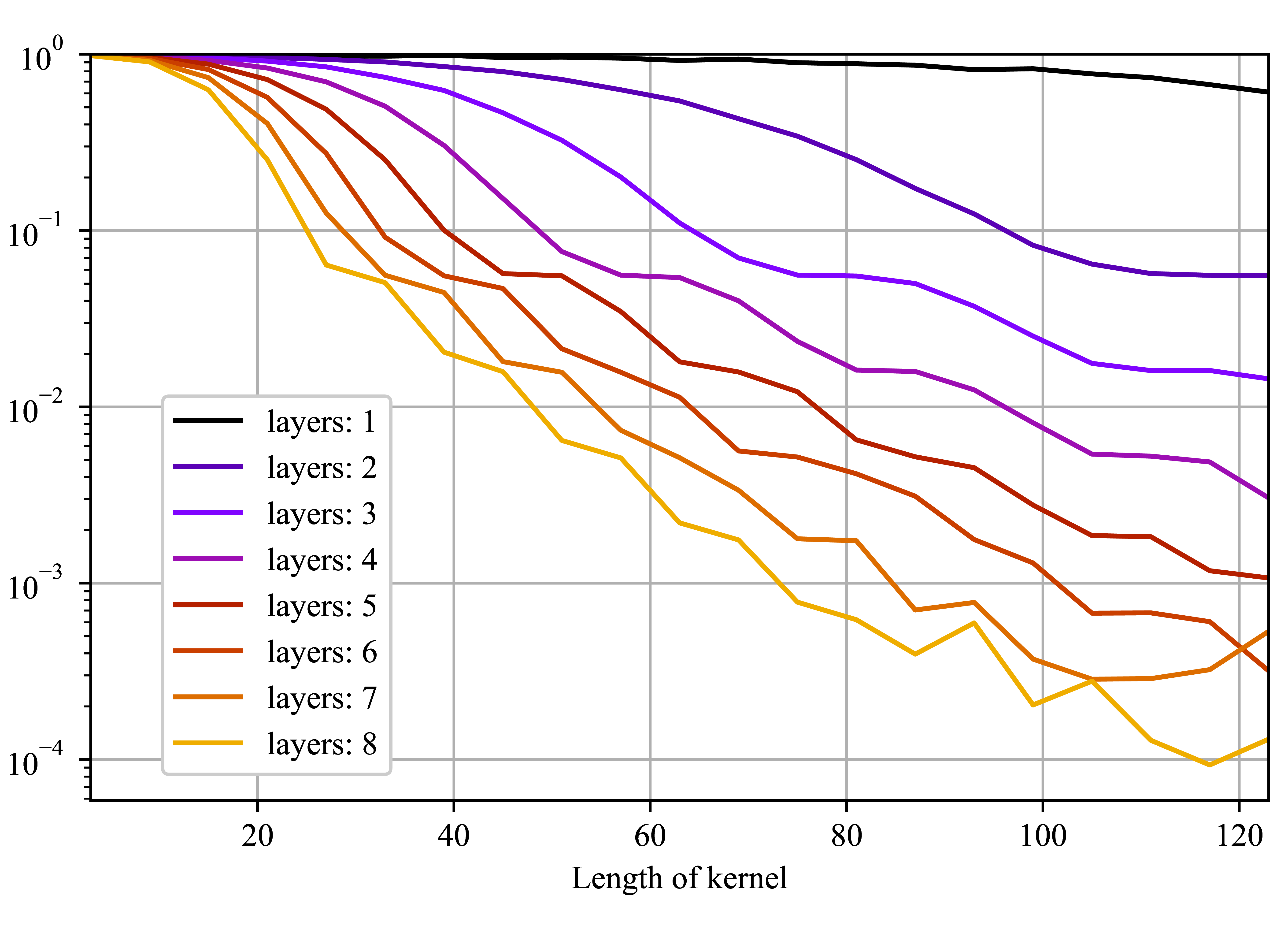}
\caption{
Loss of the CCNN the spectrum learning task for different number of layers as a function of the kernel size.
}
\label{fig:kernel-sweep}
\end{figure}

Figure \ref{fig:kernel-sweep} confirms our assumption that the number of layers influences the complexity of the optimization task, as a single layer CCNN struggles to tackle the spectral learning task.
Furthermore, we can see that the higher the number of layers, the faster the model improves as the number of kernels is increased.

Furthermore, even for a low loss, the filter of the CCNN may still diverge from the LS-FIR.
For example we can examine the case for a frequency peak at 23 Hz, with a kernel size of 75 with 6 layers on Figure \ref{fig:out_of_phase}.
This filter is able to filter the correct and bandwidth and is symmetrical.
However, it is out of phase with the LS-FIR.

\begin{figure}[ht]
\centerline{\includegraphics[width=0.8\columnwidth]{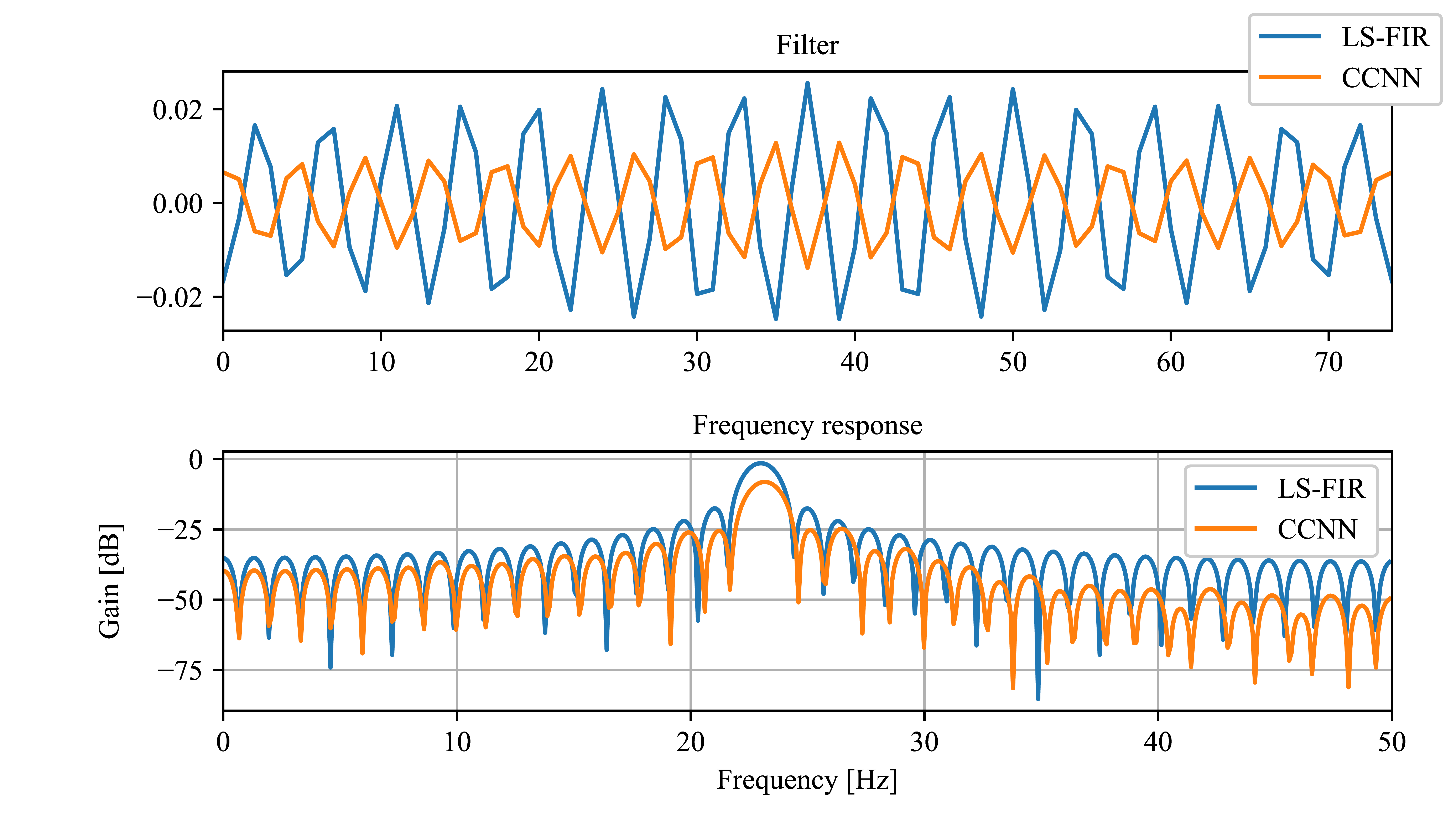}}
\caption{
A comparison of the weights and spectrum the LS-FIR and CCNN filters for a band-pass peak at 23 Hz.
The CCNN here is composed of 6 layers with a kernel width of 75.
}
\label{fig:out_of_phase}
\end{figure}

Finally, we perform a similar parameter sweep using tanh activations and compare their approximation with linear CNNs.
We show two sweeps side by side on Figure \ref{fig:linear_v_tanh}.
Overall, the CCNN with tanh activations is able to better fit the data with a lower number of parameters.

\begin{figure}[h]
    \centering
    \subfloat[Linear activation sweep.]{%
        \includegraphics[width=0.5\columnwidth]{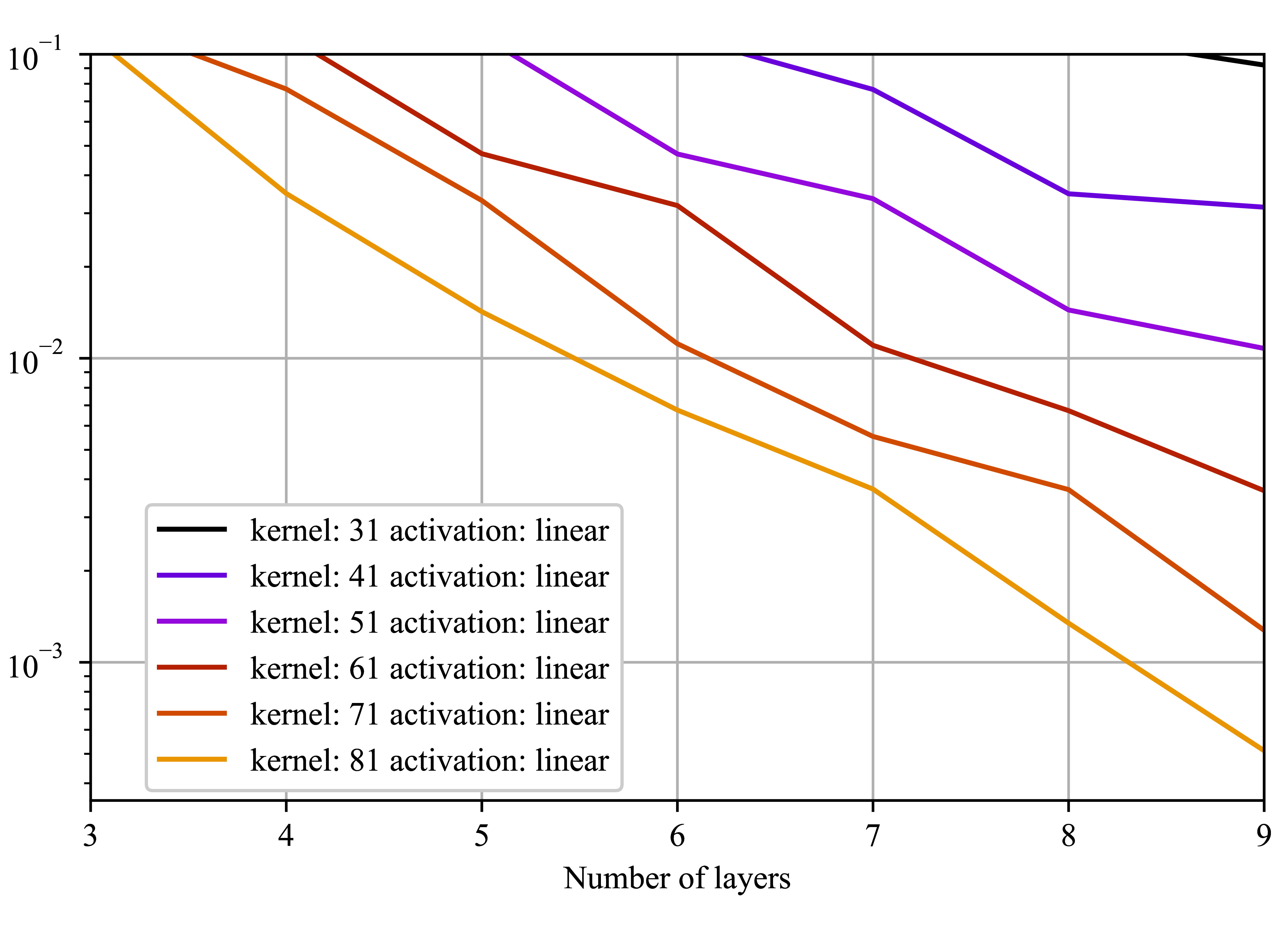}%
    }
    \hfill
    \subfloat[Tanh activation sweep.]{%
        \includegraphics[width=0.5\columnwidth]{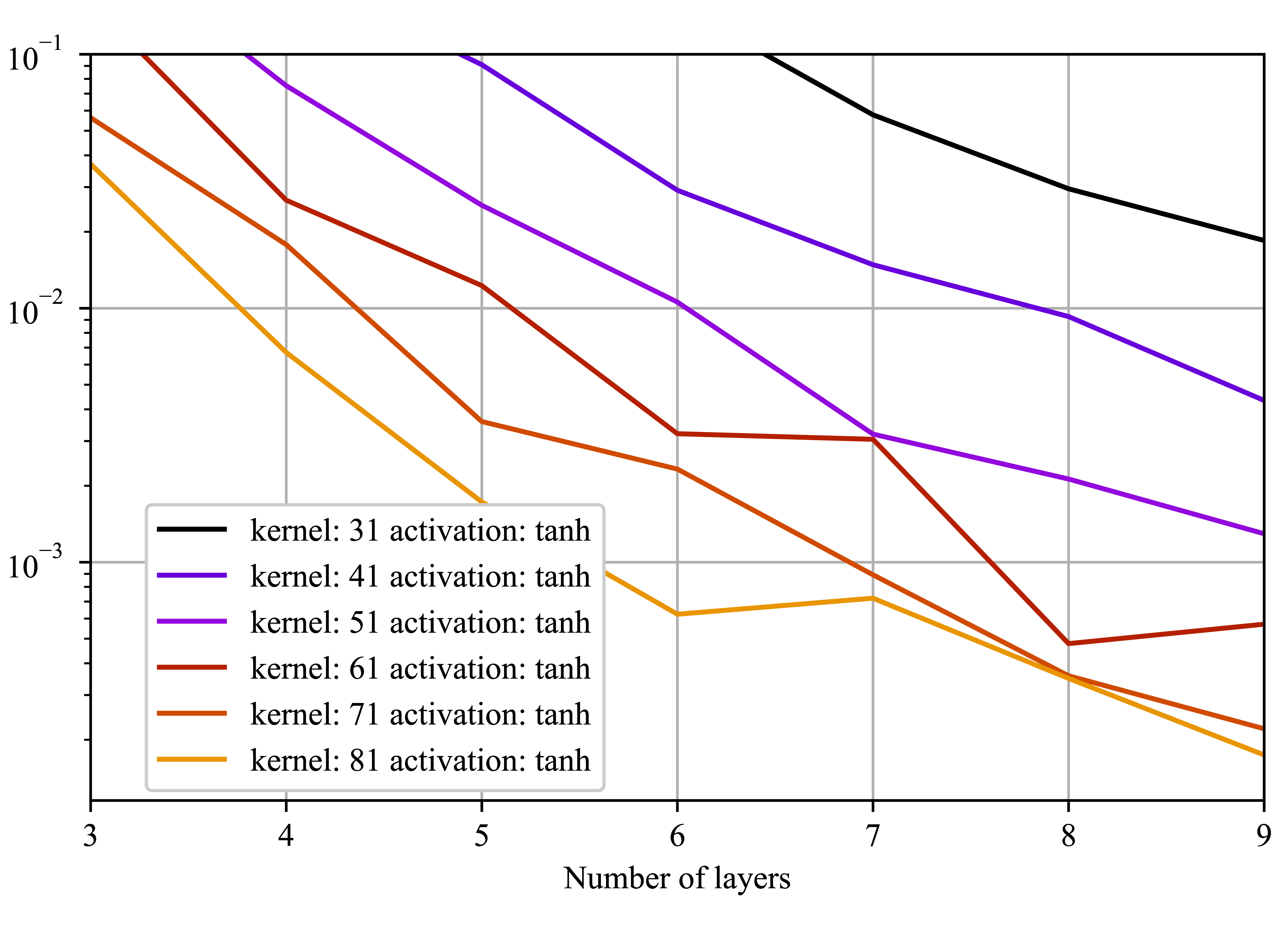}%
    }
    \caption{
        Hyperparameter sweeps for the CCNN with linear and tanh activations.
    }
    \label{fig:linear_v_tanh}
\end{figure}
}

\bibliographystyle{IEEEtranN}
\bibliography{bibliography}

\begin{thebibliography}{67}
\providecommand{\natexlab}[1]{#1}
\providecommand{\url}[1]{#1}
\csname url@samestyle\endcsname
\providecommand{\newblock}{\relax}
\providecommand{\bibinfo}[2]{#2}
\providecommand{\BIBentrySTDinterwordspacing}{\spaceskip=0pt\relax}
\providecommand{\BIBentryALTinterwordstretchfactor}{4}
\providecommand{\BIBentryALTinterwordspacing}{\spaceskip=\fontdimen2\font plus
\BIBentryALTinterwordstretchfactor\fontdimen3\font minus \fontdimen4\font\relax}
\providecommand{\BIBforeignlanguage}[2]{{%
\expandafter\ifx\csname l@#1\endcsname\relax
\typeout{** WARNING: IEEEtranN.bst: No hyphenation pattern has been}%
\typeout{** loaded for the language `#1'. Using the pattern for}%
\typeout{** the default language instead.}%
\else
\language=\csname l@#1\endcsname
\fi
#2}}
\providecommand{\BIBdecl}{\relax}
\BIBdecl

\bibitem[Alaa and van~der Schaar(2019)]{Alaa2019}
\BIBentryALTinterwordspacing
A.~M. Alaa and M.~van~der Schaar, ``Demystifying black-box models with symbolic metamodels,'' in \emph{Advances in Neural Information Processing Systems}, H.~Wallach, H.~Larochelle, A.~Beygelzimer, F.~d\textquotesingle Alch\'{e}-Buc, E.~Fox, and R.~Garnett, Eds., vol.~32.\hskip 1em plus 0.5em minus 0.4em\relax Curran Associates, Inc., 2019. [Online]. Available: \url{https://proceedings.neurips.cc/paper_files/paper/2019/file/567b8f5f423af15818a068235807edc0-Paper.pdf}
\BIBentrySTDinterwordspacing

\bibitem[Alzubaidi et~al.(2021)Alzubaidi, Zhang, Humaidi, Al-Dujaili, Duan, Al-Shamma, Santamaría, Fadhel, Al-Amidie, and Farhan]{Alzubaidi2021:a}
\BIBentryALTinterwordspacing
L.~Alzubaidi, J.~Zhang, A.~J. Humaidi, A.~Al-Dujaili, Y.~Duan, O.~Al-Shamma, J.~Santamaría, M.~A. Fadhel, M.~Al-Amidie, and L.~Farhan, ``Review of deep learning: concepts, cnn architectures, challenges, applications, future directions,'' \emph{Journal of Big Data}, vol.~8, no.~1, p.~53, 2021. [Online]. Available: \url{https://doi.org/10.1186/s40537-021-00444-8}
\BIBentrySTDinterwordspacing

\bibitem[Karniadakis et~al.(2021)Karniadakis, Kevredis, Lu, Perdikaris, Wang, and Yang]{Karniadakis2021}
G.~E. Karniadakis, I.~G. Kevredis, L.~Lu, P.~Perdikaris, S.~Wang, and L.~Yang, ``Physics-informed machine learning,'' \emph{Nature Reviews Physics}, vol.~3, pp. 422--440, 2021.

\bibitem[Reynders and Roeck(2014)]{Reynders2014:b}
\BIBentryALTinterwordspacing
E.~Reynders and G.~D. Roeck, ``Operational modal analysis in civil engineering: An overview,'' in \emph{Encyclopedia of Earthquake Engineering}.\hskip 1em plus 0.5em minus 0.4em\relax Springer, 2014, pp. 1--9, living reference work entry, first published online: 01 January 2014. [Online]. Available: \url{https://doi.org/10.1007/978-3-642-36197-5_338-1}
\BIBentrySTDinterwordspacing

\bibitem[Bayo et~al.(1989)Bayo, Papadopoulos, Stubbe, and Serna]{Bayo1989}
\BIBentryALTinterwordspacing
E.~Bayo, P.~Papadopoulos, J.~Stubbe, and M.~A. Serna, ``Inverse dynamics and kinematics of multi- link elastic robots: An iterative frequency domain approach,'' \emph{The International Journal of Robotics Research}, vol.~8, no.~6, pp. 49--62, 1989. [Online]. Available: \url{https://doi.org/10.1177/027836498900800604}
\BIBentrySTDinterwordspacing

\bibitem[Boiko(2008)]{Boiko2008}
I.~Boiko, \emph{Discontinuous control systems: frequency-domain analysis and design}.\hskip 1em plus 0.5em minus 0.4em\relax Springer Science \& Business Media, 2008.

\bibitem[Strachan(2004)]{Strachan2004}
A.~Strachan, ``Normal modes and frequencies from covariances in molecular dynamics or monte carlo simulations,'' \emph{The Journal of chemical physics}, vol. 120, no.~1, pp. 1--4, 2004.

\bibitem[Reynders(2021)]{Reynders2012}
E.~Reynders, ``System identification methods for (operational) modal analysis: Review and comparison,'' \emph{Archives of Computational Methods in Engineering}, vol.~19, pp. 51--124, 2021.

\bibitem[Kerschen et~al.(2009)Kerschen, Peeters, Golinval, and Vakakis]{Kerschen2009}
G.~Kerschen, M.~Peeters, J.~C. Golinval, and A.~F. Vakakis, ``Nonlinear normal modes, part i: A useful framework for the structural dynamicist,'' \emph{Mechanical Systems and Signal Processing}, vol.~23, no.~1, pp. 170--194, 2009.

\bibitem[Belouchrani et~al.(1997)Belouchrani, Abed-Meraim, Cardoso, and Moulines]{Belouchrani1997}
A.~Belouchrani, K.~Abed-Meraim, J.-F. Cardoso, and E.~Moulines, ``A blind source separation technique using second-order statistics,'' \emph{IEEE Transactions on Signal Processing}, vol.~45, no.~2, pp. 434--444, 1997.

\bibitem[Kerschen et~al.(2007)Kerschen, Poncelet, and Golinval]{Kerschen2007}
G.~Kerschen, F.~Poncelet, and J.-C. Golinval, ``Physical interpretation of independent component analysis in structural dynamics,'' \emph{Mechanical Systems and Signal Processing}, vol.~21, no.~4, pp. 1561--1575, 2007.

\bibitem[Yang and Nagarajaiah(2013)]{Yang2013}
Y.~Yang and S.~Nagarajaiah, ``Time-frequency blind source separation using independent component analysis for output-only modal identification of highly damped structures,'' \emph{Journal of Structural Engineering}, vol. 139, no.~10, pp. 1780--1793, 2013.

\bibitem[Nagarajaiah and Yang(2017)]{Nagarajaiah2017}
S.~Nagarajaiah and Y.~Yang, ``Modeling and harnessing sparse and low-rank data structure: A new paradigm for structural dynamics, identification, damage detection, and health monitoring,'' \emph{Structural Control and Health Monitoring}, vol.~24, no.~1, 2017.

\bibitem[Reynders et~al.(2014)Reynders, De~Roeck, Beer, Kougioumtzoglou, Patelli, and Au]{Reynders:2014a}
E.~Reynders, G.~De~Roeck, M.~Beer, I.~Kougioumtzoglou, E.~Patelli, and I.-K. Au, ``\BIBforeignlanguage{eng}{Vibration-based damage identification: the z24 benchmark},'' pp. 1--8, 2014.

\bibitem[Shih et~al.(2011)Shih, Thambiratnam, and Chan]{Shih2011}
H.~W. Shih, D.~P. Thambiratnam, and T.~H.~T. Chan, ``Damage detection in truss bridges using vibration based multi‐criteria approach,'' \emph{Structural Engineering and Mechanics}, vol.~39, no.~2, pp. 187--206, 2011.

\bibitem[Jang et~al.(2012)Jang, Spencer, and Sim]{Jang2012}
S.~A. Jang, B.~F. Spencer, and S.~H. Sim, ``A decentralized receptance‐based damage detection strategy for wireless smart sensors,'' \emph{Smart Materials and Structures}, vol.~21, no.~5, pp. 1--12, 2012.

\bibitem[Nguyen et~al.(2016)Nguyen, Chan, and Thambiratnam]{Nguyen2016}
K.~D. Nguyen, T.~H.~T. Chan, and D.~P. Thambiratnam, ``Structural damage identification based on change in geometric modal strain energy‐eigenvalue ratio,'' \emph{Smart Materials and Structures}, vol.~25, no.~7, p. 075032, 2016.

\bibitem[Spiridonakos et~al.(2016)Spiridonakos, Chatzi, and Sudret]{Spiridonakos:2016a}
\BIBentryALTinterwordspacing
M.~D. Spiridonakos, E.~N. Chatzi, and B.~Sudret, ``Polynomial chaos expansion models for the monitoring of structures under operational variability,'' \emph{ASCE-ASME Journal of Risk and Uncertainty in Engineering Systems, Part A: Civil Engineering}, vol.~2, no.~3, p. B4016003, 2016. [Online]. Available: \url{https://ascelibrary.org/doi/abs/10.1061/AJRUA6.0000872}
\BIBentrySTDinterwordspacing

\bibitem[Shokrani et~al.(2016)Shokrani, Dertimanis, Chatzi, and Savoia]{Shokrani2016}
Y.~Shokrani, V.~K. Dertimanis, E.~N. Chatzi, and M.~Savoia, ``Structural damage localization under varying environmental conditions,'' in \emph{11th HSTAM International Congress on Mechanics}, Athens, Greece, 2016.

\bibitem[Loh et~al.(2015)Loh, Hung, Chen, and Hsu]{Loh2015}
C.~H. Loh, T.~Y. Hung, S.~F. Chen, and W.~T. Hsu, ``Damage detection in bridge structure using vibration data under random travelling vehicle loads,'' \emph{Journal of Physics: Conference Series}, vol. 628, p. 012044, 2015.

\bibitem[K.(2016)]{YSSD2016}
Y.~S. S. D.~L. K., ``Application of wavelet transform in structural health monitoring,'' Master's Theses, Western Michigan University, 2016, master of Science in Engineering.

\bibitem[Ding and Chen(2013)]{Ding2013}
K.~Ding and T.~P. Chen, ``Study on damage detection of bridge based on wavelet multi-scale analysis,'' \emph{Advanced Materials Research}, vol. 640, no.~1, pp. 1010--1014, 2013.

\bibitem[Hester and Gonzalez(2012)]{Hester2012}
D.~Hester and A.~Gonzalez, ``A wavelet-based damage detection algorithm based on bridge acceleration response to a vehicle,'' \emph{Mechanical Systems and Signal Processing}, vol.~28, pp. 145--166, 2012.

\bibitem[Mcgetrick and Kim(2012)]{Mcgetrick2012}
P.~J. Mcgetrick and C.~W. Kim, ``Wavelet based damage detection approach for bridge structures utilising vehicle vibration,'' in \emph{Proceedings of the 9th German Japanese Bridge Symposium (GJBS09)}, Kyoto, Japan, 2012.

\bibitem[Aguirre et~al.(2013)Aguirre, Gaviria, and Montejo]{Aguirre2013}
D.~A. Aguirre, C.~A. Gaviria, and L.~A. Montejo, ``Wavelet‐based damage detection in reinforced concrete structures subjected to seismic excitations,'' \emph{Journal of Earthquake Engineering}, vol.~17, no.~8, pp. 1103--1125, 2013.

\bibitem[Morales-Valdez et~al.(2020)Morales-Valdez, Lopez-Pacheco, and Yu]{MoralesValdez2020}
J.~Morales-Valdez, M.~Lopez-Pacheco, and W.~Yu, ``Automated damage location for building structures using the hysteretic model and frequency domain neural networks,'' \emph{Structural Control and Health Monitoring}, June 2020, cited by 8.

\bibitem[He et~al.(2021)He, Chen, Liu, and Zhang]{He2021}
Y.~He, H.~Chen, D.~Liu, and L.~Zhang, ``A framework of structural damage detection for civil structures using fast fourier transform and deep convolutional neural networks,'' \emph{Applied Sciences}, vol.~11, no.~19, p. 9345, 2021, submission received: 12 September 2021; Revised: 2 October 2021; Accepted: 3 October 2021; Published: 8 October 2021.

\bibitem[Jian et~al.(2024)Jian, Xia, Duthé, Bacsa, Liu, and Chatzi]{Jian:2024a}
\BIBentryALTinterwordspacing
X.~Jian, Y.~Xia, G.~Duthé, K.~Bacsa, W.~Liu, and E.~Chatzi, ``Using graph neural networks and frequency domain data for automated operational modal analysis of populations of structures,'' 2024. [Online]. Available: \url{https://arxiv.org/abs/2407.06492}
\BIBentrySTDinterwordspacing

\bibitem[Nguyen et~al.(2024)Nguyen, Salamak, Katunin, Poprawa, and Gerges]{Nguyen2024}
D.~C. Nguyen, M.~Salamak, A.~Katunin, G.~Poprawa, and M.~Gerges, ``Vibration-based shm of railway steel arch bridge with orbit-shaped image and wavelet-integrated cnn classification,'' \emph{Engineering Structures}, vol. 315, p. 118431, September 2024.

\bibitem[{van der Schaaf} and {van Hateren}(1996)]{VanDerSchaf:1996a}
\BIBentryALTinterwordspacing
A.~{van der Schaaf} and J.~{van Hateren}, ``Modelling the power spectra of natural images: Statistics and information,'' \emph{Vision Research}, vol.~36, no.~17, pp. 2759--2770, 1996. [Online]. Available: \url{https://www.sciencedirect.com/science/article/pii/0042698996000028}
\BIBentrySTDinterwordspacing

\bibitem[Owen et~al.(2001)Owen, Eccles, Choo, and Woodings]{Owen2001}
\BIBentryALTinterwordspacing
J.~Owen, B.~Eccles, B.~Choo, and M.~Woodings, ``The application of auto–regressive time series modelling for the time–frequency analysis of civil engineering structures,'' \emph{Engineering Structures}, vol.~23, no.~5, pp. 521--536, 2001. [Online]. Available: \url{https://www.sciencedirect.com/science/article/pii/S0141029600000596}
\BIBentrySTDinterwordspacing

\bibitem[Zhang et~al.(2022)Zhang, Zhou, Wen, and Sun]{Zhang2022}
\BIBentryALTinterwordspacing
C.~Zhang, T.~Zhou, Q.~Wen, and L.~Sun, ``Tfad: A decomposition time series anomaly detection architecture with time-frequency analysis,'' in \emph{Proceedings of the 31st ACM International Conference on Information \& Knowledge Management}, ser. CIKM '22.\hskip 1em plus 0.5em minus 0.4em\relax New York, NY, USA: Association for Computing Machinery, 2022, p. 2497–2507. [Online]. Available: \url{https://doi.org/10.1145/3511808.3557470}
\BIBentrySTDinterwordspacing

\bibitem[Rahaman et~al.(2019)Rahaman, Baratin, Arpit, Lin, Hamprecht, Bengio, and Courville]{Rahaman2019}
N.~Rahaman, A.~Baratin, D.~Arpit, F.~D.~M. Lin, F.~Hamprecht, Y.~Bengio, and A.~Courville, ``On the spectral bias of neural networks,'' in \emph{Proceedings of the 36th International Conference on Machine Learning, PMLR}, K.~Chaudhuri and R.~Salakhutdinov, Eds.\hskip 1em plus 0.5em minus 0.4em\relax Long Beach, California: Curran Associates, Inc., 2019, pp. 5301--5310.

\bibitem[Jacot et~al.(2021)Jacot, Gabiel, and Hongler]{Jacot2021}
A.~Jacot, F.~Gabiel, and C.~Hongler, ``Neural tangent kernel: convergence and generalization in neural networks,'' in \emph{Proceedings of the 53rd Annual ACM SIGACT Symposium on Theory of Computing}, S.~Khuller, Ed.\hskip 1em plus 0.5em minus 0.4em\relax New York NY, United States: Association for Computing Machinery, 2021, p.~6.

\bibitem[Stankovic and Mandic(2023)]{Stankovic2023}
L.~Stankovic and D.~Mandic, ``Convolutional neural networks demystified: A matched filtering perspective-based tutorial,'' \emph{IEEE Transactions on Systems, Man, and Cybernetics: Systems}, vol.~53, no.~6, pp. 3614--3628, 2023.

\bibitem[Rumelhart and McClelland(1987)]{Rumelhart1987}
D.~E. Rumelhart and J.~L. McClelland, Eds., \emph{Parallel Distributed Processing: Explorations in the Microstructure of Cognition: Foundations}.\hskip 1em plus 0.5em minus 0.4em\relax Cambridge, MA: MIT Press, 1987.

\bibitem[Hochreiter and Schmidhuber(1997)]{Hochreiter:1997a}
S.~Hochreiter and J.~Schmidhuber, ``Long short-term memory,'' \emph{Neural Computation}, vol.~9, no.~8, pp. 1735--1780, 1997.

\bibitem[Vaswani et~al.(2017)Vaswani, Shazeer, Parmar, Uszkoreit, Jones, Gomez, Kaiser, and Polosukhin]{Vaswani2017}
A.~Vaswani, N.~Shazeer, N.~Parmar, J.~Uszkoreit, L.~Jones, A.~N. Gomez, L.~Kaiser, and I.~Polosukhin, ``Attention is all you need,'' in \emph{Advances in Neural Information Processing Systems}, I.~Guyon, U.~V. Luxburg, S.~Bengio, H.~Wallach, R.~Fergus, S.~Vishwanathan, and R.~Garnett, Eds., San Diego, CA, 2017.

\bibitem[Soderstrom and Stoica(1988)]{Soderstrom1988}
T.~Soderstrom and P.~Stoica, ``On some system identification techniques for adaptive filtering,'' \emph{IEEE Transactions on Circuits and Systems}, vol.~35, no.~4, pp. 457--461, 1988.

\bibitem[Gunasekar et~al.(2018)Gunasekar, Lee, Soudry, and Srebro]{Gunasekar:2018a}
\BIBentryALTinterwordspacing
S.~Gunasekar, J.~D. Lee, D.~Soudry, and N.~Srebro, ``Implicit bias of gradient descent on linear convolutional networks,'' in \emph{Advances in Neural Information Processing Systems}, S.~Bengio, H.~Wallach, H.~Larochelle, K.~Grauman, N.~Cesa-Bianchi, and R.~Garnett, Eds., vol.~31.\hskip 1em plus 0.5em minus 0.4em\relax Curran Associates, Inc., 2018. [Online]. Available: \url{https://proceedings.neurips.cc/paper_files/paper/2018/file/0e98aeeb54acf612b9eb4e48a269814c-Paper.pdf}
\BIBentrySTDinterwordspacing

\bibitem[{De Ryck} et~al.(2021){De Ryck}, Lanthaler, and Mishra]{DeRyck:2021a}
\BIBentryALTinterwordspacing
T.~{De Ryck}, S.~Lanthaler, and S.~Mishra, ``On the approximation of functions by tanh neural networks,'' \emph{Neural Networks}, vol. 143, pp. 732--750, 2021. [Online]. Available: \url{https://www.sciencedirect.com/science/article/pii/S0893608021003208}
\BIBentrySTDinterwordspacing

\bibitem[Kovacevic et~al.(2013)Kovacevic, Goyal, and Vetterli]{Kovacevic2013}
J.~Kovacevic, V.~K. Goyal, and M.~Vetterli, ``Fourier and wavelet signal processing,'' \emph{Fourier Wavelets. org}, pp. 1--294, 2013.

\bibitem[Zhang et~al.(2020)Zhang, Xu, Yang, Chen, Zhou, Liu, Li, Lin, and Ying]{Zhang2020}
X.~Zhang, J.~Xu, J.~Yang, L.~Chen, H.~Zhou, X.~Liu, H.~Li, T.~Lin, and Y.~Ying, ``Understanding the learning mechanism of convolutional neural networks in spectral analysis,'' \emph{Analytica Chimia Acta}, vol. 1119, pp. 41--51, 2020.

\bibitem[Li et~al.(2021)Li, Kovachki, Azizzadenesheli, liu, Bhattacharya, Stuart, and Anandkumar]{li2021fourier}
\BIBentryALTinterwordspacing
Z.~Li, N.~B. Kovachki, K.~Azizzadenesheli, B.~liu, K.~Bhattacharya, A.~Stuart, and A.~Anandkumar, ``Fourier neural operator for parametric partial differential equations,'' in \emph{International Conference on Learning Representations}, 2021. [Online]. Available: \url{https://openreview.net/forum?id=c8P9NQVtmnO}
\BIBentrySTDinterwordspacing

\bibitem[Chen and Billings(1992)]{Chen1992}
S.~Chen and S.~A. Billings, ``Neural networks for nonlinear dynamic system modelling and identification,'' \emph{International Journal of control}, vol.~56, no.~2, pp. 319--346, 1992.

\bibitem[Chance et~al.(1998)Chance, Worden, and Tomlinson]{Chance1998}
\BIBentryALTinterwordspacing
J.~Chance, K.~Worden, and G.~Tomlinson, ``Frequency domain analysis of narx neural networks,'' \emph{Journal of Sound and Vibration}, vol. 213, no.~5, pp. 915--941, 1998. [Online]. Available: \url{https://www.sciencedirect.com/science/article/pii/S0022460X98915395}
\BIBentrySTDinterwordspacing

\bibitem[Gonzalez and Yu(2018)]{Gonzalez2018}
J.~Gonzalez and W.~Yu, ``Non-linear system modeling using lstm neural networks,'' \emph{IFAC-PapersOnLine}, vol.~51, no.~13, pp. 485--489, 2018.

\bibitem[Piga et~al.(2021)Piga, Forgione, and Mejari]{Piga2021}
D.~Piga, M.~Forgione, and M.~Mejari, ``Deep learning with transfer functions: new applications in system identification,'' \emph{IFAC-PapersOnLine}, vol.~54, no.~7, pp. 415--420, 2021.

\bibitem[Andersson et~al.(2019)Andersson, Ribeiro, Tiels, Wahlström, and Schön]{Andersson2019}
C.~Andersson, A.~H. Ribeiro, K.~Tiels, N.~Wahlström, and T.~B. Schön, ``Deep convolutional networks in system identification,'' in \emph{2019 IEEE 58th Conference on Decision and Control (CDC)}, A.~G. Aghdam, Ed.\hskip 1em plus 0.5em minus 0.4em\relax Nice, France: IEEE, 2019, pp. 3670--3676.

\bibitem[Gu et~al.(2022)Gu, Goel, and R\'e]{Gu2022}
A.~Gu, K.~Goel, and C.~R\'e, ``Efficiently modeling long sequences with structured state spaces,'' in \emph{The International Conference on Learning Representations ({ICLR})}, 2022.

\bibitem[Gu and Dao(2023)]{Gu2023}
A.~Gu and T.~Dao, ``Mamba: Linear-time sequence modeling with selective state spaces,'' 2023.

\bibitem[Li et~al.(2024)Li, Li, Wang, He, Wang, Wang, and Qiao]{Li2024}
K.~Li, X.~Li, Y.~Wang, Y.~He, Y.~Wang, L.~Wang, and Y.~Qiao, ``Videomamba: State space model for efficient video understanding,'' 2024.

\bibitem[Zhu et~al.(2024)Zhu, Liao, Zhang, Wang, Liu, and Wang]{Zhu2024}
L.~Zhu, B.~Liao, Q.~Zhang, X.~Wang, W.~Liu, and X.~Wang, ``Vision mamba: Efficient visual representation learning with bidirectional state space model,'' \emph{arXiv preprint arXiv:2401.09417}, 2024.

\bibitem[Parks and McClellan(1972)]{Parks1972}
T.~Parks and J.~McClellan, ``Chebyshev approximation for nonrecursive digital filters with linear phase,'' \emph{IEEE Transactions on Circuit Theory}, vol.~19, no.~2, pp. 189--194, 1972.

\bibitem[Bendat and Piersol(2010)]{Bendat2010}
J.~S. Bendat and A.~G. Piersol, Eds., \emph{Random Data: Analysis and Measurement Procedures}.\hskip 1em plus 0.5em minus 0.4em\relax New York NY, United States: John Wiley and Sons, Inc., 2010.

\bibitem[Paszke et~al.(2019)Paszke, Gross, Massa, Lerer, Bradbury, Chanan, Killeen, Lin, Gimelshein, Antiga, Desmaison, Kopf, Yang, DeVito, Raison, Tejani, Chilamkurthy, Steiner, Fang, Bai, and Chintala]{pytorch}
\BIBentryALTinterwordspacing
A.~Paszke, S.~Gross, F.~Massa, A.~Lerer, J.~Bradbury, G.~Chanan, T.~Killeen, Z.~Lin, N.~Gimelshein, L.~Antiga, A.~Desmaison, A.~Kopf, E.~Yang, Z.~DeVito, M.~Raison, A.~Tejani, S.~Chilamkurthy, B.~Steiner, L.~Fang, J.~Bai, and S.~Chintala, ``Pytorch: An imperative style, high-performance deep learning library,'' in \emph{Advances in Neural Information Processing Systems 32}, H.~Wallach, H.~Larochelle, A.~Beygelzimer, F.~d\textquotesingle Alch\'{e}-Buc, E.~Fox, and R.~Garnett, Eds.\hskip 1em plus 0.5em minus 0.4em\relax Curran Associates, Inc., 2019, pp. 8024--8035. [Online]. Available: \url{http://papers.neurips.cc/paper/9015-pytorch-an-imperative-style-high-performance-deep-learning-library.pdf}
\BIBentrySTDinterwordspacing

\bibitem[Glorot and Bengio(2010)]{Glorot2010}
X.~Glorot and Y.~Bengio, ``Understanding the difficulty of training deep feedforward neural networks,'' \emph{Proceedings of Machine Learning Research}, vol.~30, no.~9, pp. 249--256, 2010.

\bibitem[Virtanen et~al.(2020)Virtanen, Gommers, Oliphant, Haberland, Reddy, Cournapeau, Burovski, Peterson, Weckesser, Bright, {van der Walt}, Brett, Wilson, Millman, Mayorov, Nelson, Jones, Kern, Larson, Carey, Polat, Feng, Moore, {VanderPlas}, Laxalde, Perktold, Cimrman, Henriksen, Quintero, Harris, Archibald, Ribeiro, Pedregosa, {van Mulbregt}, and {SciPy 1.0 Contributors}]{scipy2020}
P.~Virtanen, R.~Gommers, T.~E. Oliphant, M.~Haberland, T.~Reddy, D.~Cournapeau, E.~Burovski, P.~Peterson, W.~Weckesser, J.~Bright, S.~J. {van der Walt}, M.~Brett, J.~Wilson, K.~J. Millman, N.~Mayorov, A.~R.~J. Nelson, E.~Jones, R.~Kern, E.~Larson, C.~J. Carey, {\.I}.~Polat, Y.~Feng, E.~W. Moore, J.~{VanderPlas}, D.~Laxalde, J.~Perktold, R.~Cimrman, I.~Henriksen, E.~A. Quintero, C.~R. Harris, A.~M. Archibald, A.~H. Ribeiro, F.~Pedregosa, P.~{van Mulbregt}, and {SciPy 1.0 Contributors}, ``{{SciPy} 1.0: Fundamental Algorithms for Scientific Computing in Python},'' \emph{Nature Methods}, vol.~17, pp. 261--272, 2020.

\bibitem[Hornik et~al.(1989)Hornik, Stinchcombe, and White]{Hornik:1989a}
\BIBentryALTinterwordspacing
K.~Hornik, M.~Stinchcombe, and H.~White, ``Multilayer feedforward networks are universal approximators,'' \emph{Neural Networks}, vol.~2, no.~5, pp. 359--366, 1989. [Online]. Available: \url{https://www.sciencedirect.com/science/article/pii/0893608089900208}
\BIBentrySTDinterwordspacing

\bibitem[Cybenko(1989)]{Cybenko:1989a}
G.~Cybenko, ``Approximation by superpositions of a sigmoidal function,'' \emph{Mathematics of control, signals and systems}, vol.~2, no.~4, pp. 303--314, 1989.

\bibitem[Bianchini and Scarselli(2014)]{Bianchini:2014a}
M.~Bianchini and F.~Scarselli, ``On the complexity of neural network classifiers: A comparison between shallow and deep architectures,'' \emph{IEEE Transactions on Neural Networks and Learning Systems}, vol.~25, no.~8, pp. 1553--1565, 2014.

\bibitem[Kingma and Welling(2013)]{Kingma:2013a}
\BIBentryALTinterwordspacing
D.~P. Kingma and M.~Welling, ``Auto-encoding variational bayes,'' 2013. [Online]. Available: \url{https://arxiv.org/abs/1312.6114}
\BIBentrySTDinterwordspacing

\bibitem[Maeck and De~Roeck(2003)]{Maeck2003}
J.~Maeck and G.~De~Roeck, ``Description of z24 benchmark,'' \emph{Mechanical Systems and Signal Processing}, vol.~17, no.~1, pp. 127--131, 2003.

\bibitem[Brincker et~al.(2000)Brincker, Zhang, and Andersen]{FDD2000}
R.~Brincker, L.~Zhang, and P.~Andersen, ``Modal identification from ambient responses using frequency domain decomposition,'' \emph{Proceedings of the International Modal Analysis Conference - IMAC}, vol.~1, 01 2000.

\bibitem[Savitzky and Golay(1964)]{Savitzky1964}
A.~Savitzky and M.~J.~E. Golay, ``Smoothing and differentiation of data by simplified least squares procedures,'' \emph{Analytical Chemistry}, vol.~36, no.~8, pp. 1627--1639, 1964.

\bibitem[Selesnick(2005)]{Selesnick2005}
I.~Selesnick, ``Linear-phase fir filter design by least-squares,'' \emph{Connexions}, 2005.

\bibitem[Heinig and Rost(2011)]{Heining2011}
G.~Heinig and K.~Rost, ``Fast algorithms for toeplitz and hankel matrices,'' \emph{Linear Algebra and its Applications}, vol. 435, no.~1, pp. 1--59, 2011.

\end{thebibliography}

\end{document}